\documentclass[10pt,twocolumn,letterpaper]{article}

\usepackage[pagenumbers]{cvpr} 

\usepackage{amsmath} 
\newcommand{\overtext}[1]{\ensuremath{\overline{\text{#1}}}}
\definecolor{trf}{rgb}{0.5, 0.5, 1}
\definecolor{livingCoral}{HTML}{F36F63}

\usepackage{multirow}
\usepackage{makecell}
\usepackage{colortbl}
\definecolor{LightYellow}{rgb}{1.0, 1.0, 0.8}
\definecolor{LightRed}{rgb}{1.0, 0.8, 0.8}

\definecolor{bestpink}{RGB}{255,179,186}
\definecolor{secondpeach}{RGB}{255,219,172}

\newcommand{\Bs}[1]{\textbf{#1}}
\newcommand{\Ss}[1]{\textit{#1}}
\newcommand{\PB}{\bfseries}
\newcommand{\PS}{\itshape}

\definecolor{cvprblue}{rgb}{0.21,0.49,0.74}
\usepackage[pagebackref,breaklinks,colorlinks,allcolors=cvprblue]{hyperref}

\def\paperID{41443} 
\def\confName{CVPR}
\def\confYear{2026}

\title{ArtPro: Self-Supervised Articulated Object Reconstruction with Adaptive Integration of Mobility Proposals}

\author{
Xuelu Li$^{1*}$ \quad
Zhaonan Wang$^{1*}$ \quad
Xiaogang Wang$^{2\dagger}$ \quad
Lei Wu$^{1}$ \quad
Manyi Li$^{1\dagger}$ \quad
Changhe Tu$^{1}$\\[0.5em]
$^{1}$Shandong University \qquad $^{2}$Southwest University\\[0.3em]
{\tt\footnotesize \{xtluli, d\_wzn\}@mail.sdu.edu.cn \quad wangxiaogang@swu.edu.cn \quad \{i\_lily, manyili, chtu\}@sdu.edu.cn}
}

\begin{document}
\maketitle
\let\thefootnote\relax\footnotetext{$^{*}$Equal contribution. $^{\dagger}$Corresponding author.}
\begin{abstract}

Reconstructing articulated objects into high-fidelity digital twins is crucial for applications such as robotic manipulation and interactive simulation. 
Recent self-supervised methods using differentiable rendering frameworks like 3D Gaussian Splatting remain highly sensitive to the initial part segmentation. Their reliance on heuristic clustering or pre-trained models often causes optimization to converge to local minima, especially for complex multi-part objects.
To address these limitations, we propose ArtPro, a novel self-supervised framework that introduces adaptive integration of mobility proposals. Our approach begins with an over-segmentation initialization guided by geometry features and motion priors, generating part proposals with plausible motion hypotheses. During optimization, we dynamically merge these proposals by analyzing motion consistency among spatial neighbors, while a collision-aware motion pruning mechanism prevents erroneous kinematic estimation. Extensive experiments on both synthetic and real-world objects demonstrate that ArtPro achieves robust reconstruction of complex multi-part objects, significantly outperforming existing methods in accuracy and stability.



\end{abstract}    
\section{Introduction}
\label{sec:intro}
Articulated objects such as cabinets, laptops, and scissors are ubiquitous in our daily lives. Creating their high-fidelity digital twins, including accurate reconstructions of part geometry, appearance, and kinematic structure, is the cornerstone for enabling robotic manipulation, virtual reality, and interactive scene simulation. However, this task remains highly challenging, as it requires jointly solving a set of intertwined sub-problems: part segmentation, motion parameter estimation, as well as high-quality appearance and geometry reconstruction.

\begin{figure}
  \centering
  \includegraphics[width=\linewidth]{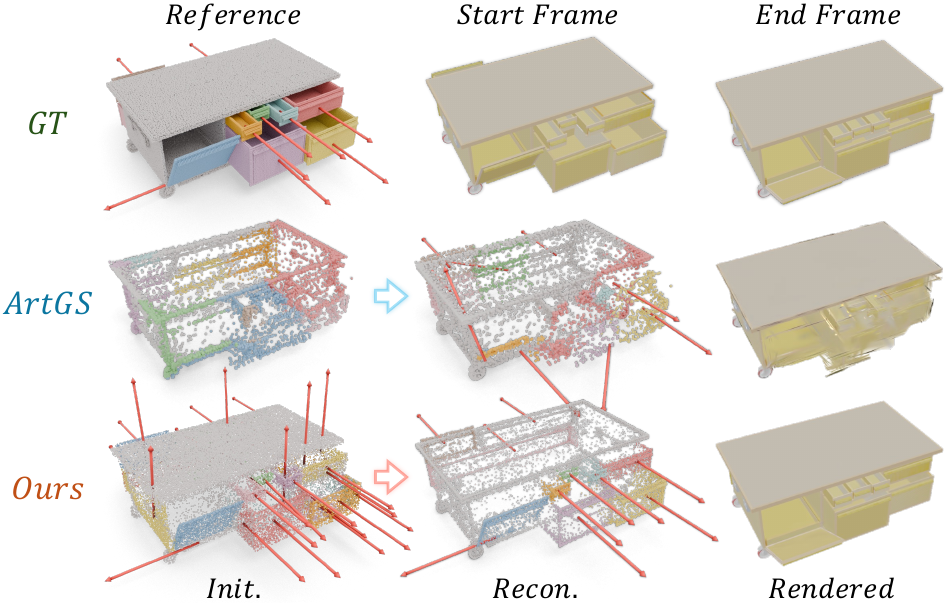}
  \caption{Existing methods like ArtGS~\cite{liu2025artgs} are highly sensitive to initial part segmentation, leading to inaccurate motion and geometry. Our method, ArtPro, leverages a prior-guided mobility initialization and adaptively merges mobility proposals during optimization, achieving robust reconstruction of complex multi-part articulated objects.}
  \label{fig:teaser}
  \vspace{-10pt}
\end{figure}

Existing reconstruction approaches largely fall into two paradigms. On one hand, the feed-forward reconstruction methods~\cite{kawana2023detection, chen2024urdformer, dai2024automated, gao2025partrm, wu2025dipo, su2025artformer, qiu2025articulate} leverage powerful data-driven priors from diffusion or vision-language models to directly predict articulated structures from inputs like images or text, achieving remarkable inference speed. However, their generality is achieved either through coarse geometric abstractions or by being constrained to their training data. As a result, they often fail on unseen object categories or produce low-fidelity reconstructions, thus lacking instance-specific details. On the other hand, per-instance optimization methods~\cite{mu2021sdf, wu2022d, deng2024articulate, wu2025reartgs, liu2025videoartgs}, as represented by frameworks like NeRF or 3D Gaussian Splatting (3DGS), directly optimize the reconstruction for a specific instance without relying on those pre-trained models, achieving superior fidelity. Yet, they possess a critical weakness, i.e. \textbf{extreme sensitivity to the initialization of movable parts}. As shown in Figure 1, an initial segmentation from heuristic clustering that fails on complex multi-part objects can still yield a small rendering loss, causing the optimization to converge to an incorrect kinematic structure (a local minimum). This strong dependency is particularly detrimental for objects with multiple movable parts, which severely hinders reconstruction efficiency and limits their practical utility in real-world applications.

To overcome these limitations, we introduce \textbf{ArtPro}, a robust and self-supervised framework for articulated object reconstruction based on 3DGS. Our core idea is to forgo the brittle ``guess-the-segmentation-once" paradigm and instead adopt a more intelligent and robust \textbf{``propose-verify-merge''} pipeline. That is, we reframe part segmentation not as a fixed starting point, but as a dynamic process that is continuously refined during optimization.

The robustness and efficacy of ArtPro are driven by three key designs that form a self-correcting optimization system:
\begin{itemize}
	\item \textbf{Prior-Guided Mobility Proposal Initialization}: 
    We generate over-segmented mobility proposals by integrating geometry features with motion priors, establishing a robust initialization for subsequent optimization.
	
	\item \textbf{Adaptive Proposal Integration}: 
    We dynamically merge mobility part proposals during the 3DGS optimization process by assessing motion consistency between adjacent regions, merging those that produce similar rendered results under exchanged motion parameters. 
	
	\item \textbf{Collision-Aware Motion Pruning}: 
    We introduce an active motion pruning mechanism that monitors inter-part collisions during optimization and calibrates problematic motion parameters to prevent local minima and ensure stable kinematic estimation.
\end{itemize}

Across extensive synthetic and real-world experiments, ArtPro demonstrates strong robustness and accuracy, consistently generating digital twins with precise  geometry and kinematics. The method excels particularly on complex objects with multiple adjacent parts, significantly outperforming existing state-of-the-art alternatives.

\section{Related Works}
\label{sec:related_work}

\subsection{Understanding of 3D Articulated Objects}

Prior research on articulated object understanding aims to infer part-level mobility and kinematic structures from sensory inputs, spanning several key directions. 
A substantial body of work focuses on \emph{detecting movable parts}, either from single static images~\cite{jiang2022opd, sun2023opdmulti} or by discovering articulation in videos~\cite{qian2022understanding}. 
Another major direction is \emph{joint parameter estimation}, which recovers kinematic parameters such as joint axes, pivot point, and articulation angles. 
Further extending this line of work, methods have been developed for various inputs: some use point clouds~\cite{wang2019shape2motion, yan2020rpm, li2020category, jiang2022ditto, 2023Category, fu2024capt}, others rely on RGB-D images from single or multiple views~\cite{hu2017learning, abbatematteo2019learning, jain2021screwnet, jain2022distributional, liu2022toward, che2024op}, and some leverage videos~\cite{liu2020nothing} or 4D dynamic point clouds~\cite{liu2023building}.

To overcome the limitations of passive perception, which often suffer from ambiguous or insufficient viewpoints, active perception methods have been developed. These methods employ strategies that enable robots to actively interact with the environment or plan optimal viewing trajectories, thereby disambiguating the articulated structure and facilitating a more accurate estimation of its kinematic parameters~\cite{yan2023interaction, zeng2024mars, bhuniainteractive,zhang2025iaao}. More recently, the field has increasingly leveraged the powerful reasoning capabilities of Vision-Language Models (VLMs) for articulated object understanding. By fine-tuning VLMs on these tasks, researchers have enabled them to directly estimate joint parameters and perform kinematic reasoning from a combination of visual and textual inputs~\cite{huang2024a3vlm, vora2025articulate, qiu2025articulate}. However, despite producing shapes with plausible structures and joint parameters, these methods rely heavily on object datasets with extensive part-level annotations and consequently exhibit limited generalization ability.


\subsection{Reconstruction of 3D Articulated Objects}

Current methodologies for 3D articulated object reconstruction fall into two paradigms: \emph{feed-forward} reconstruction and \emph{per-instance optimization}, each presenting a unique set of trade-offs between efficiency, generality, and fidelity.

The first paradigm employs \emph{feed-forward} models to directly predict articulated structures from inputs such as single/multiple images~\cite{kawana2023detection, chen2024urdformer, dai2024automated, liu2024singapo, gao2025partrm, zhang2021strobenet, patil2023rosi, heppert2023carto, mandi2024real2code, wu2025dipo}, text prompts~\cite{su2025artformer, qiu2025articulate}, or articulation graphs~\cite{liu2024cage}. Leveraging the powerful data-driven priors from diffusion models~\cite{lei2023nap, gao2025meshart, luo2025physpart} or vision-language models~\cite{le2024articulate, qiu2025articulate, vora2025articulate}, these methods achieve remarkable inference speed. However, this efficiency comes at a significant cost. To ensure generality, they often rely on the coarse geometric abstractions, such as bounding boxes, primitive shapes, or parts retrieved from limited databases, which fundamentally limit their ability to reconstruct fine-grained, instance-specific geometry and high-fidelity appearances. Therefore, their output quality is bounded by the scope of their training data and the simplicity of their output representations.

In contrast, the second paradigm employs \emph{per-instance optimization}, bypassing pre-trained feedforward models to directly optimize a reconstruction for a specific object instance. The populer strategy is to rely on implicit representations, Neural Radiance Fields (NeRF)~\cite{mu2021sdf, wu2022d, deng2024articulate} and, more recently, 3D Gaussian Splatting~\cite{wu2025reartgs, liu2025artgs}. However, the scalability of these methods is severely limited by their strong dependency on the initial part segmentation. This is particularly problematic for complex multi-part objects, where an imperfect initialization often causes the optimization to converge to incorrect kinematic structures, making them largely ineffective in practice. 
To address these limitations, we introduce ArtPro, a novel framework designed to overcome the aforementioned challenges by adaptively initializing and integrating mobility proposals.

\begin{figure*}
  \centering
  \includegraphics[width=\linewidth]{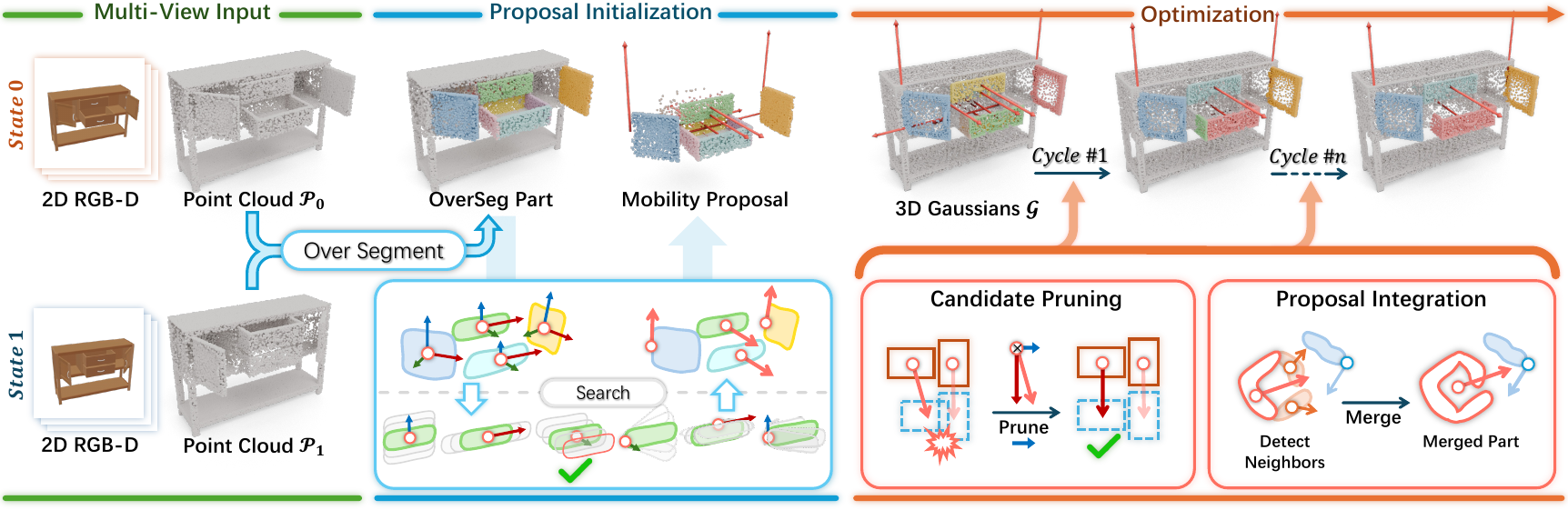}
  \caption{\emph{ArtPro} reconstructs articulated objects from multi-view RGBD images of two states. The pipeline begins by initializing part-motion proposals through over-segmentation and a mixed-variable search. These proposals are then refined via a self-supervised optimization that updates the parts and their motion parameters using transformable 3D Gaussians. The optimization incorporates motion pruning and proposal integration operations to calibrate motions and merge proposals into coherent movable parts. Finally, a post-processing refinement stabilizes the appearance, geometry, and motion parameters of the reconstruction.}
  \label{fig:pipeline}
\end{figure*}

\section{Method}
\label{sec:methods}

\subsection{Problem Formulation}
\label{problem}

Our goal is to reconstruct realistic digital twins of articulated objects from captured multi-view images. As shown in Figure~\ref{fig:pipeline}, the input comprises two motion states of an articulated object, each state represented by a set of RGBD images. We parameterize the object states using a continuous variable $t$, and consider the two given states as the start ($t=0$) and end ($t=1$), respectively. Thus, any $t\in(0,1)$ represents an intermediate state. Assuming the object contains one static part and $M$ movable parts, the output of our approach consists of the segmented parts and their motion parameters, enabling realistic motion simulation of the reconstructed object.


The segmented parts are encoded using the 3D Gaussian Splatting (3DGS) representation. Specifically, in addition to the vanilla Gaussian parameters, we define part-aware probability fields and assign probability values for the Gaussian centered at $x_i$: $P_s(x_i)$ indicating its likelihood of belonging to the static part, and $\{P_m(x_i)\}^M_{m=1}$ for the movable parts. In this way, we assign the Gaussians to different parts based on the conditional probability
\begin{equation}
\begin{aligned}
\hat{P}_m(x_i)&=\big(1-P_s(x_i)\big)\frac{P_m(x_i)}{\sum^M_{k=1}P_k(x_i)},\\
    \mathcal{G}_m&=\Big\{g_i\in\mathcal{G}\big|\hat{P}_m(x_i)>\epsilon\Big\},
\end{aligned}
\end{equation}
where $\mathcal G_m$ denotes the $m$-th movable part, $\epsilon=0.01$ is the threshold, $x_i$ is the center of Gaussian $g_i\in\mathcal{G}$. The union of all the Gaussians $\mathcal G$ represents the whole object.

The part motion parameters are defined according to the joint types. For prismatic joints, the motion of the movable part is parameterized by a translation vector $t_m\in\mathcal R^3$. For revolute joints, the motion parameter is defined as a 6DoF rotation $r_m\in \mathcal R^6$~\cite{zhou2019continuity} and a center position $c_m\in \mathcal R^3$. Put them together, we can apply the rigid transformation of part $m$ to a Gaussian $g_i$ centered at $x_i$:
\begin{equation}
    \hat x_i=R_m(x_i-c_m)+c_m+t_m,\quad \hat q_i=R_m\otimes q_i,
\end{equation}
where $R_m$ is the rotation matrix of $r_m$ and $q_i$ is the quaternion parameter of the Gaussian $g_i$.



With the above representation, we can apply the motion parameters to transform the articulated object $\mathcal G$
\begin{equation}
    \hat{\mathcal G}=\mathcal{T}(\mathcal G).
\end{equation}
To render the transformed articulated object $\hat{\mathcal G}$ with 3DGS differentiable rasterizer, we assign the adjusted opacity values $\hat \alpha_i=\hat{P}_m(x_i) \alpha_i$ and $\hat \alpha_i=P_s(x_i) \alpha_i$ to the Gaussians of movable parts and static part, respectively. This allows us to optimize the parts and their motion parameters of the articulated object based on the multi-view RGBD images.

\subsection{Mobility Proposal Initialization}

Since each part is attached with its own motion parameters, the part initialization significantly impacts subsequent optimization. To achieve robust reconstruction of complex multi-part objects, our strategy is to initialize with an over-segmentation of mobility proposals, which are then progressively merged during the optimization process.




The process begins with the over-segmentation of point clouds $\mathcal{P}^t = \{\mathcal{P}^0, \mathcal{P}^1\}$, which are constructed from input multi-view depth maps. The movable set $\hat{\mathcal{P}}_0$ is obtained by selecting points in $\mathcal{P}^0$ whose nearest-neighbor distance to $\mathcal{P}^1$ exceeds a threshold $\tau$. Next, $n$ seed points are selected from $\hat{\mathcal{P}}_0$ via farthest point sampling. We then utilize point-wise features from a pre-trained segmentation model~\cite{liu2025partfield} to grow $n$ over-segmented parts around these seeds. Finally, we merge parts with more than $80\%$ overlap to produce $M$ movable part proposals $\{\hat{\mathcal{P}}_m\}$. Note that the proposal count $M$ does not necessarily equal to the ground-truth part number of the object.



We initialize the motion parameters for these proposals based on the heuristic that joint axes are typically perpendicular to part surfaces or aligned with their edges. For each part $\mathcal{P}_m$, we extract its oriented bounding box (OBB) and three principal component axes $\{\mathcal{X}_m, \mathcal{Y}_m, \mathcal{Z}_m\}$. We then select the most reliable axis to initialize the motion parameters by solving a mixed-variable optimization problem:
\begin{equation}
\arg\min_{\hat{a}, \phi, d} \ \text{CD}\left( R_m(\hat{a}, \phi)(\hat{\mathcal{P}}_m - c) + c + d \cdot \hat{a}\rightarrow \mathcal{P}^1 \right),
\end{equation}
where we jointly solve for the discrete joint axis choice $\hat{a} \in \{\mathcal{X}_m, \mathcal{Y}_m, \mathcal{Z}_m\}$, the rotation angle $\phi$, and the translation magnitude $d$. Here, $R_m(\hat{a}, \phi)$ denotes the rotation about axis $\hat{a}$ by angle $\phi$, and $c$ is the center of the OBB face closest to $\mathcal{P}^1$. The objective is to minimize the one-sided Chamfer distance $\text{CD}(\rightarrow)$ between the transformed movable parts and the target point cloud $\mathcal{P}^1$. The variables are constrained to $d \in [-0.5, 0.5]$ meters and $\phi \in [-80^\circ, 80^\circ]$.

After obtaining the optimal solution, the motion parameters are initialized to an intermediate state as
\begin{equation}
R_m \leftarrow R_m(\hat{a}, \phi/2), \quad t_m \leftarrow (d/2) \cdot \hat{a}, \quad c_m \leftarrow c,
\end{equation}
which helps stabilize the subsequent optimization by reducing the transformation magnitude to half.

\subsection{Adaptive Proposal Integration Optimization}
\label{sec:adaptive_algorithm}

We propose an adaptive optimization algorithm to progressively merge initial mobility proposals, ultimately determining the coherent movable parts of the articulated object. As in Figure~\ref{fig:pipeline}, the algorithm begins by establishing associations between the Gaussian primitives and the initialized proposals. It then refines these proposals through an adaptive optimization cycle, which consists of three key components: (1) \emph{Gaussian primitive optimization}, which is the main body of the cycle, refines the segmentation and motion parameters of the proposals by iteratively updating the Gaussian primitives. (2) \emph{Motion pruning operation} calibrates the motion parameters of the part proposals based on their motion trajectories. (3) \emph{Proposal integration operation} merges pairs of proposals that belong to the same part and updates the associated Gaussian primitives. The cycle repeats until no proposals can be merged.



\noindent\textbf{Gaussian-Proposal Association.} We initialize two sets of Gaussian primitives from the point clouds $\mathcal P^0$ and $\mathcal P^1$. Following~\cite{liu2025artgs}, we assign the part-assignment probability for each Gaussian primitive using a Gaussian Mixture Model (GMM). Specifically, the region of the $m$-th movable part proposal is represented by a Gaussian component $\mathcal{N}(\mu_m,\Sigma_m)$ with a positive mixture weight $\omega_m \in \mathbb{R}^+$. $\mu_m$ and $\Sigma_m$ are initialized from the mean and variance of the point set $\hat{\mathcal{P}}_m$. The likelihood of a Gaussian primitive located at $x$ belonging to movable part $m$ is then given by
\begin{equation}
    P_m(x) = \omega_m \cdot \mathcal{N}(x; \mu_m, \Sigma_m).
\end{equation}
And that of the static part $P_s(x)$ is initialized as zero. 
This probabilistic association allows us to assign the Gaussians to the parts as described in Section~\ref{problem} and transform Gaussian primitives according to the motion parameters.



\noindent\textbf{Gaussian Primitive Optimization.} The goal of this optimization is to iteratively refine the parameters of Gaussian primitives and their associated proposal motion parameters. In each iteration, we transform the Gaussians from the start state to the end state as described in Section~\ref{problem} and render them with the differentiable rasterizer. Therefore, we obtain the Gaussian sets $\mathcal{G}$, their centers $\{x_i\}$, and the rendered images of the transformed articulated object.

We define a series of loss terms as the optimization objectives. The RGBD image loss $\mathcal{L}_{I}$ of end state and one-sided chamfer distance loss $\mathcal{L}_{cd}$ encourage the transformed Gaussians to align with the underlying surface of the end state, in order to ensure the reconstruction fidelity and enable the following proposal integration by checking the spatial relation of Gaussian centers. The part contrastive loss $\mathcal{L}_{pc}$ encourages the assignment of each Gaussian to a single dominant part by suppressing the probabilities of all non-maximum part proposals. The local smoothness loss $\mathcal{L}_{ls}$ enforces spatial smoothness on the static part probability field by encouraging similar values for neighboring Gaussian primitives. The regularization term $\mathcal{L}_{reg}$ encourages the compactness of each part by regularizing the part probabilities to align with a Gaussian distribution.

Finally, the objective function of the optimization is
\begin{equation}
\mathcal{L}=\mathcal{L}_{I}+\lambda_{cd}\mathcal{L}_{cd}+\lambda_{pc}\mathcal{L}_{pc}+\lambda_{ls}\mathcal{L}_{ls}+\lambda_{reg}\mathcal{L}_{reg}.
\end{equation}
The detailed formulation of each loss term is presented in the supplementary material.

\noindent\textbf{Motion Pruning Strategy.}
We calibrate the joint axes every 100 iterations during the Gaussian primitive optimization by measuring the overlap volume between the proposal pairs along their motion trajectories. It prevents the optimization from becoming stuck in local minima.

Specifically, for parts $i$ and $j$, we compute their oriented bounding boxes (OBBs) in both the original (i.e. \(b^0_i, b^0_j\)) and transformed states (i.e. \(b^1_i, b^1_j\)). The overlap volumes between the two parts in these states are calculated as \(v_{i,j}^0 = \text{Vol}(b^0_i \cap b^0_j)\) and \(v_{i,j}^1 = \text{Vol}(b^1_i \cap b^1_j)\). A collision is flagged if the overlap volume of the end state is larger than that of the start state, i.e. $\Delta v_{i,j}=v_{i,j}^1-v_{i,j}^0>\tau_v$ with $\tau_v=10^{-4}$. 
Once a collision is detected, the colliding axis \(a_{\text{col}}\) is identified as the principal axis of the OBB that exhibits the largest projection of the relative translation vector between the two parts. The motion parameters are then calibrated as follows to resolve the collision.

For a \emph{prismatic joint}, the component of the translation vector \(t_m\) along the colliding axis \(a_{\text{col}}\) is pruned by projecting it onto the plane orthogonal to \(a_{\text{col}}\):
\begin{equation}
    t_m \leftarrow t_m - (t_m \cdot a_{\text{col}}) \cdot a_{\text{col}}.
\end{equation}

For a \emph{revolute joint}, the proposals with near-identity rotation angles (e.g., less than \(5^\circ\)) are suppressed by resetting the rotation axis to align with the nearest OBB edge and halving the rotation angle to avoid the mutual influence between the two joint types. 

In addition, we enforce a hard constraint on the joint type after 4K iterations. Thus, we set and fix $t_m$ as zero for revolute joints and $R_m$ as identity matrix for prismatic joints.


\noindent\textbf{Proposal Integration Operation.}
We identify and merge proposal pairs that are both spatially adjacent and kinematically consistent. This is conducted every 5K iterations of the Gaussian primitive optimization.

The spatial adjacency between two proposals, \( \mathcal{G}_i \) and \( \mathcal{G}_j \), is determined by performing the \( k \)-nearest neighbor search (with \( k = 8 \)) over the union of all movable Gaussian centers \( \cup\{\mathcal{G}_m\} \). Specifically, for each point \( x \) in \( \mathcal{G}_i \), if any of its eight nearest neighbors belongs to a different movable part \( \mathcal{G}_j \), then \( \mathcal{G}_i \) and \( \mathcal{G}_j \) are considered adjacent.

To evaluate motion consistency between an adjacent pair \( (\mathcal{G}_i, \mathcal{G}_j) \), we define a score function based on the L1 distance between rendered depth maps. The score \( S(\mathcal{T}) \) for a given motion parameters \( \mathcal{T} \) is formulated as the sum of L1 errors across all training views \( v \) of the end state:
\begin{equation}
S(\mathcal{T}) = \sum_{\mathcal{D}_v \in \{\mathcal{D}^1_i\}} | D_v(\mathcal{T}(\mathcal{G})) - \mathcal{D}_v |, 
\end{equation}
where \( D_v(\mathcal{T}(\mathcal{G})) \) is the rendered depth map from viewpoint \( v \), and \( \mathcal{D}_v \) is the corresponding ground-truth depth map at the end state (\( t = 1 \)). For a pair \( (\mathcal{G}_i, \mathcal{G}_j) \), we compute two scores: the score with their current motion parameters, \(S(\mathcal{T})\), and the score when the motion of part $i$ is replaced by that of part $j$, denoted as \(S(\mathcal{T}^{(j\rightarrow i)})\). 
The integration criterion is that their score variation falls below a predefined threshold \( \tau \), i.e., \( |S(\mathcal{T}^{(i)}_i) - S(\mathcal{T}^{(j)}_i)| < \tau_{merge}=10^{-3} \). If a part \( \mathcal{G}_i \) has multiple adjacent parts eligible for merging, it is merged with the neighbor proposal \( \mathcal{G}_j \) for which the score variation is the smallest. This strategy ensures that we integrate only spatially adjacent parts with empirically consistent motions, effectively preventing spurious merges caused by minor rendering score jitter.

When integrating two part proposals,  we merge $\mathcal{G}_i$ and $\mathcal{G}_j$, and use the union of their Gaussian centers to initialize the updated OBB and part probability field $P_{new}(x)$. The motion of the integrated part is updated as that of $\mathcal{G}_j$.

\subsection{Post-Processing Refinement}
After repeating the adaptive optimization cycle process several times to progressively integrate the proposals into movable parts, we conduct the Gaussian primitive optimization again as the post-processing refinement to obtain the accurate reconstruction. We follow the procedures of the Gaussian primitive optimization as described in Section~\ref{sec:adaptive_algorithm} with a modified objective function.

The objective function includes the RGBD loss at both states $t=\{0,1\}$. And we introduce a collision loss \(\mathcal{L}_{col}\) to avoid the collision  between the movable and static parts of transformed object $\mathcal{T}(\mathcal{G})$. Thus the objective function is
\begin{equation}
    \mathcal{L}=\mathcal{\mathcal{L}}_{I}(\mathcal{G})+\mathcal{\mathcal{L}}_{I}(\mathcal{T}(\mathcal{G}))+\lambda_c\mathcal{\mathcal{L}}_{col}
\end{equation}
Detailed formulations are in the supplementary material.



\section{Experiments}
\label{sec:exps}


\begin{table*} 
  \centering
  \footnotesize 
  \setlength{\tabcolsep}{3pt} 
  \caption{Quantitative evaluation on our dataset. Lower($\downarrow$) is better on all metrics. We mark all joint type prediction errors and movable part count prediction errors with \textbf{F}. Window-103238 and Table-34610 are prismatic with no Axis Pos.}
  \label{tab:ours_cmp}
  \resizebox{\linewidth}{!}{
  \begin{tabular}{ll cccccccc c}
    \toprule
    & & \makecell{Table 34178 \\ (5 parts)} 
      & \makecell{Storage 40417 \\ (7 parts)}
      & \makecell{Window 103238 \\ (3 parts)}
      & \makecell{Table 23372 \\ (5 parts)}
      & \makecell{Storage 45759 \\ (5 parts)}
      & \makecell{Table 33116 \\ (4 parts)}
      & \makecell{Table 34610 \\ (6 parts)}
      & \makecell{Storage 47585 \\ (11 parts)}
      & \ \ \ All\ \ \ \\
    \midrule
    
    \multirow{3}{*}{\makecell{Axis \\ Ang}} 
    & DTA~\cite{weng2024dta}
        &	46.62	&	20.40	&	4.60	&	33.58	&	6.10	&	2.17	&	34.19	&	31.74	&	22.42	\\
    & ArtGS~\cite{liu2025artgs}
        &	4.64	&	F	&\bfseries	0.02	&	39.52	&	16.68	&	0.04	&	F	&	0.97	&	8.70$\ ^\text{F}$	\\
    & Ours
        &\bfseries	0.05	&\bfseries	0.07	&	0.06	&\bfseries	0.06	&\bfseries	0.03	&\bfseries	0.02	&\bfseries	0.09	&\bfseries	0.16	&\bfseries	0.07	\\
    \midrule

    \multirow{3}{*}{\makecell{Axis \\ Pos}} 
    & DTA~\cite{weng2024dta}
        &	21.68	&	3.37	&	-	&	2.90	&	0.66	&	0.01	&	-	&	1.15	&	4.96	\\
    & ArtGS~\cite{liu2025artgs}
        &	1.42	&	F	&	-	&	1.46	&	0.05	&\bfseries	0.00	&	-	&\bfseries	0.00	&	0.32$\ ^\text{F}$	\\
    & Ours
        &\bfseries	0.00	&\bfseries	0.00	&	-	&\bfseries	0.00	&\bfseries	0.00	&\bfseries	0.00	&	-	&\bfseries	0.00	&\bfseries	0.00	\\
    \midrule

    \multirow{3}{*}{\makecell{Part \\ Motion}} 
    & DTA~\cite{weng2024dta}
        &	25.65	&	30.17	&	0.26	&	12.79	&	32.92	&	0.11	&	0.10	&	2.51	&	13.07	\\
    & ArtGS~\cite{liu2025artgs}
        &	11.42	&	F	&	0.20	&	15.21	&	13.01	&\bfseries	0.01	&	F	&	27.65	&	9.50$\ ^\text{F}$	\\
    & Ours
        &\bfseries	0.09	&\bfseries	0.08	&\bfseries	0.00	&\bfseries	0.02	&\bfseries	0.05	&	0.04	&\bfseries	0.00	&\bfseries	0.00	&\bfseries	0.04	\\
    \midrule

    \multirow{3}{*}{CD-s}
    & DTA~\cite{weng2024dta}
        &	1.38	&	4.37	&	7.52	&	1.75	&	3.70	&	1.95	&	2.89	&	5.15	&	3.59	\\
    & ArtGS~\cite{liu2025artgs}
        &	1.87	&	0.90	&	6.61	&	2.81	&	1.83	&	1.60	&	2.55	&\bfseries	2.97	&	2.64	\\
    & Ours
        &\bfseries	0.42	&\bfseries	0.36	&\bfseries	0.11	&\bfseries	0.47	&\bfseries	0.37	&\bfseries	0.65	&\bfseries	1.15	&	3.21	&\bfseries	0.84	\\
    \midrule

    \multirow{3}{*}{CD-m}
    & DTA~\cite{weng2024dta}
        &	F	&	F	&	65.60	&	F	&	F	&	0.85	&	48.06	&	F	&	38.17$\ ^\text{F}$	\\
    & ArtGS~\cite{liu2025artgs}
        &	16.91	&	1,316.00	&	330.98	&	541.48	&	29.37	&	1.33	&	253.04	&	273.40	&	345.31	\\
    & Ours
        &\bfseries	3.58	&\bfseries	0.26	&\bfseries	0.14	&\bfseries	2.20	&\bfseries	0.48	&\bfseries	0.15	&\bfseries	13.01	&\bfseries	11.11	&\bfseries	3.86	\\
    \midrule

    \multirow{3}{*}{CD-w}
    & DTA~\cite{weng2024dta}
        &	1.06	&	0.84	&	0.44	&	0.96	&	0.85	&	1.47	&	0.91	&	2.36	&	1.11	\\
    & ArtGS~\cite{liu2025artgs}
        &	2.05	&	1.00	&	0.64	&	2.69	&	1.67	&	1.97	&	2.02	&	4.08	&	2.02	\\
    & Ours
        &\bfseries	0.45	&\bfseries	0.33	&\bfseries	0.12	&\bfseries	0.41	&\bfseries	0.36	&\bfseries	0.30	&\bfseries	1.16	&\bfseries	2.13	&\bfseries	0.65	\\
    \bottomrule
  \end{tabular}
  }

\end{table*}


\begin{table}[t] 
  \centering
  \setlength{\tabcolsep}{2pt}
  \caption{ArtGS-Multi dataset~\cite{liu2025artgs} results. Lower($\downarrow$) is better on all metrics. Table-25493 is prismatic with no Axis Pos.}
  \label{tab:artgs_cmp}
  \resizebox{\linewidth}{!}{
  \begin{tabular}{ll ccccc c}
    \toprule
    & & \makecell{Table 25493 \\ (4 parts)} & \makecell{Table 31249 \\ (5 parts)} & \makecell{Storage 45503 \\ (4 parts)} & \makecell{Storage 47468 \\ (7 parts)} & \makecell{Oven 101908 \\ (4 parts)} & \ \ \ All\ \ \ \\
    \midrule
    
    \multirow{3}{*}{\makecell{Axis\\ Ang}} 
    & DTA
        &	24.35	&	20.62	&	51.18	&	19.07	&	17.83	&	26.61	\\
    & ArtGS
        &	1.16	&\bfseries	0.04	&\bfseries	0.02	&	0.14	&	0.04	&	0.28	\\
    & Ours
        &\bfseries	0.08	&\bfseries	0.04	&	0.03	&\bfseries	0.07	&\bfseries	0.01	&\bfseries	0.05	\\
    \midrule

    \multirow{3}{*}{\makecell{Axis\\ Pos}} 
    & DTA
        &	-	&	4.20	&	2.44	&	0.31	&	6.51	&	3.37	\\
    & ArtGS
        &	-	&\bfseries	0.00	&\bfseries	0.00	&	0.02	&\bfseries	0.01	&	0.01	\\
    & Ours
        &	-	&\bfseries	0.00	&\bfseries	0.00	&\bfseries	0.00	&\bfseries	0.01	&\bfseries	0.00	\\
    \midrule

    \multirow{3}{*}{\makecell{Part\\ Motion}} 
    & DTA
        &	0.12	&	30.80	&	43.77	&	10.67	&	31.80	&	23.43	\\
    & ArtGS
        &\bfseries	0.00	&\bfseries	0.01	&\bfseries	0.03	&	0.62	&	0.23	&	0.18	\\
    & Ours
        &\bfseries	0.00	&	0.02	&\bfseries	0.03	&\bfseries	0.09	&\bfseries	0.09	&\bfseries	0.05	\\
    \midrule

    \multirow{3}{*}{CD-s}
    & DTA
        &	0.59	&	1.39	&	5.74	&	0.82	&	1.17	&	1.94	\\
    & ArtGS
        &	0.74	&	1.22	&	0.75	&	0.67	&	1.08	&	0.89	\\
    & Ours
        &\bfseries	0.20	&\bfseries	0.32	&\bfseries	0.24	&\bfseries	0.18	&\bfseries	0.55	&\bfseries	0.30	\\
    \midrule

    \multirow{3}{*}{CD-m}
    & DTA
        &	104.38	&	230.38	&	246.63	&	476.91	&	359.16	&	283.49	\\
    & ArtGS
        &	3.53	&	3.09	&	0.13	&	3.70	&	0.25	&	2.14	\\
    & Ours
        &\bfseries	0.14	&\bfseries	1.88	&\bfseries	0.08	&\bfseries	2.26	&\bfseries	0.11	&\bfseries	0.89	\\
    \midrule

    \multirow{3}{*}{CD-w}
    & DTA
        &	0.55	&	1.00	&	0.88	&	0.71	&	1.01	&	0.83	\\
    & ArtGS
        &	0.74	&	1.16	&	0.88	&	0.70	&	1.03	&	0.90	\\
    & Ours
        &\bfseries	0.18	&\bfseries	0.39	&\bfseries	0.25	&\bfseries	0.17	&\bfseries	0.47	&\bfseries	0.29	\\
    \bottomrule
  \end{tabular}
  }
  \vspace{-2mm}

\end{table}

\subsection{Settings}

\noindent\textbf{Datasets.} We evaluate all the methods with articulated objects collected from the datasets of existing works~\cite{liu2023paris,liu2025artgs}, including two-part objects and multi-part objects. In addition, we select 8 multi-part objects with 3-11 parts from the PartNet-Mobility dataset~\cite{xiang2020sapien}. These articulated objects contain multiple parts with diverse motion structures and close locations, which poses particular challenges for reconstructing their accurate articulation structures. 

\noindent\textbf{Metrics}. Following the evaluation protocol of ArtGS~\cite{liu2025artgs}, we assess performance with the metrics of both mesh reconstruction and articulation estimation. For mesh reconstruction, we uniformly sample 10K points on the reconstructed mesh and the ground-truth mesh, then compute the chamfer distance for the whole object (CD-w), the static parts (CD-s), and the movable parts (CD-m). For articulation estimation, we compute the angular error (Axis Ang.) and distance (Axis Pos.) between the predicted and ground-truth joint axes, with the latter only for revolute joints. We also report part motion error (Part Motion), measuring rotational geodesic distance error (in degrees) for revolute joints and Euclidean distance error (in meters) for prismatic joints.



\subsection{Comparisons}

\subsubsection{Comparison on Two-part Objects}
\begin{figure}
  \centering
  \includegraphics[width=\linewidth]{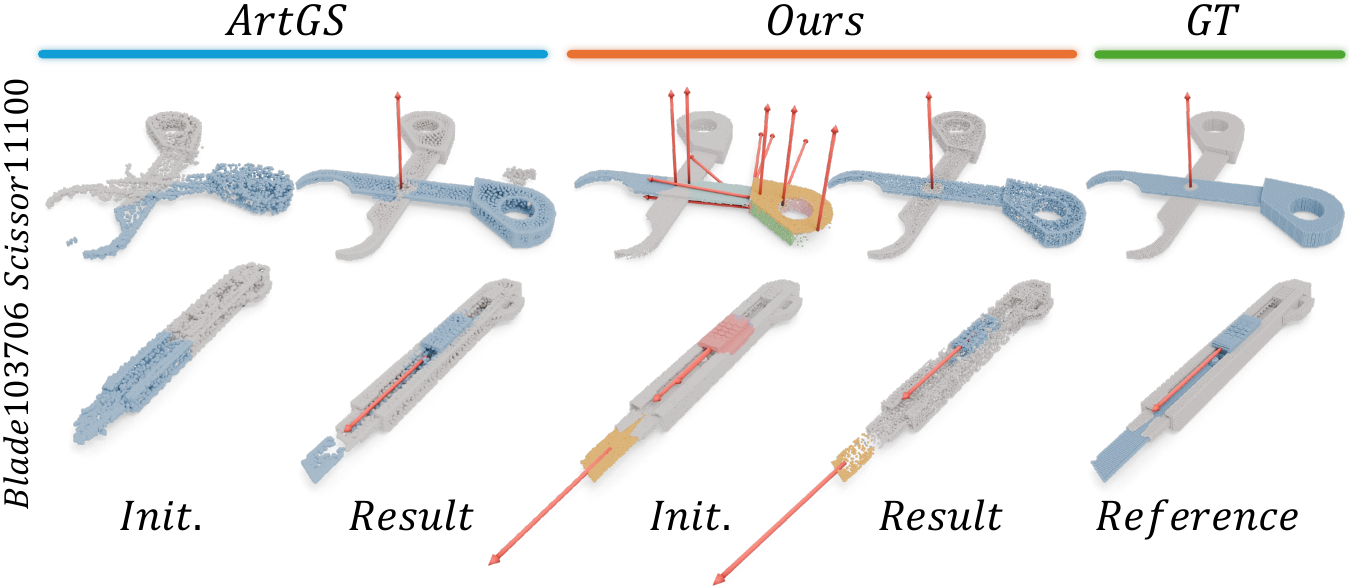}
  \caption{The part initialization and reconstruction results of ArtGS~\cite{liu2025artgs} and ours on two-part objects. Although our approach doesn't merge the two disadjacent movable parts (second row), we still obtain accurate motion and geometry reconstruction.
  }
  \label{fig:two_obj}
  \vspace{-2mm}
\end{figure}

The quantitative comparison with  PARIS~\cite{liu2023paris}, ArticulatedGS~\cite{guo2025articulatedgs}, DTA~\cite{weng2024dta}, ArtGS~\cite{liu2025artgs}, is presented in the supplementary material, showing that the 3DGS-based approaches produce competitive reconstruction results. Figure~\ref{fig:two_obj} provides the visual results of two-part objects using ArtGS~\cite{liu2025artgs} and our approach. It demonstrates that the optimization results of ArtGS (and other 3DGS-based approaches) relies heavily on the initialization. By contrast, our method adaptively merge the over-segmented part proposals during optimization, thus achieving more robust reconstruction results.

\begin{figure*}[!ht]
  \centering
  \includegraphics[width=\linewidth]{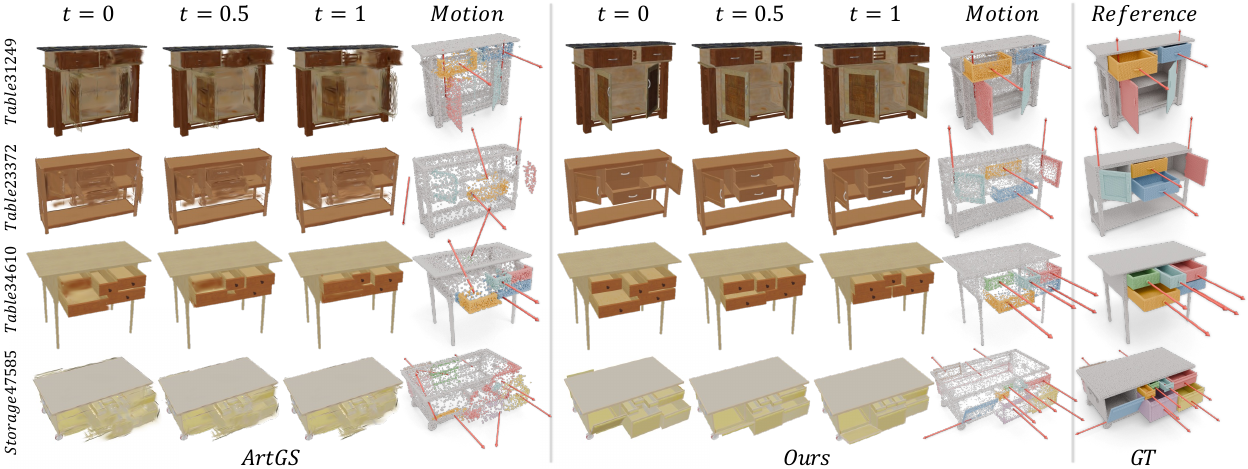}
  \caption{The reconstructed articulated objects in different motion states ($t=\{0, 0.5, 1\}$) and their part-motion structures.}
  \label{fig:multi_obj}
  \vspace{-3mm}
\end{figure*}
 
\begin{figure*}[!ht]
  \centering
  \includegraphics[width=\linewidth]{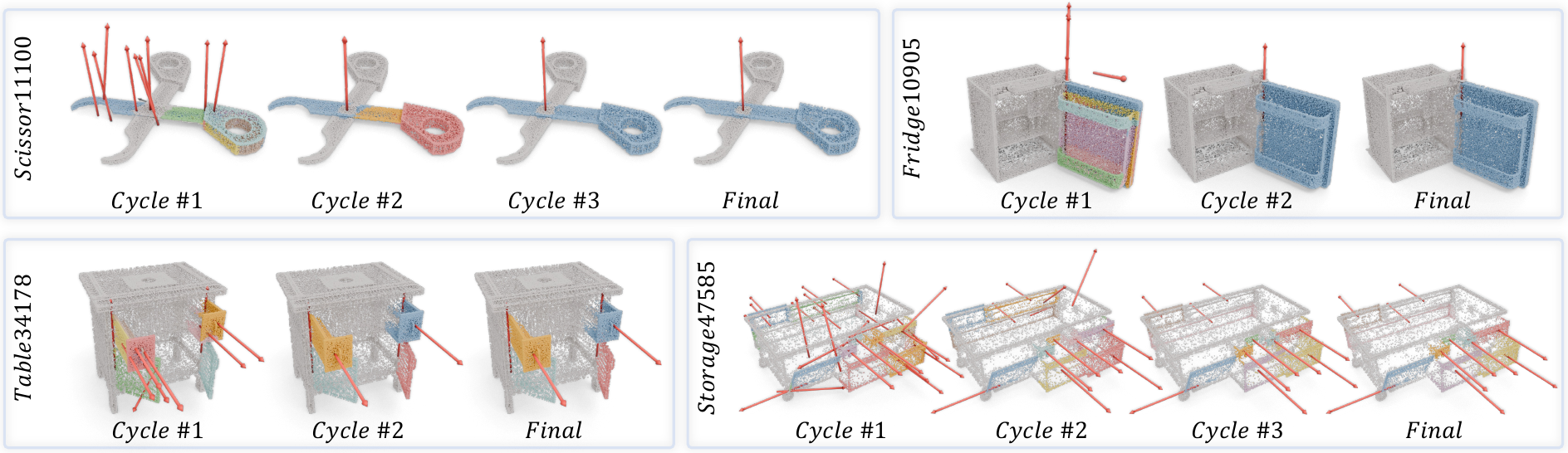}
  \caption{The intermediate results during our adaptive proposal integration optimization. We show the optimized Gaussians (with their center points) and the estimated motions before the proposal integration operation at each cycle. }
  \label{fig:merge_vis}
  \vspace{-2mm}
\end{figure*}

It is worth noting that our method doesn't necessarily merge the disconnected components together, such as the Blade-103706 shown in second row of Figure~\ref{fig:two_obj}. For this case, although the slider and blade are assigned as different parts, our method still successfully estimates their accurate motion parameters due to the adaptive proposal integration.




\subsubsection{Comparison on Multi-part Objects} 
We further evaluate the performance of multi-part object reconstruction methods, i.e. DTA~\cite{weng2024dta} and ArtGS~\cite{liu2025artgs}. Table~\ref{tab:artgs_cmp} and Table~\ref{tab:ours_cmp} reports the quantitative results. Since DTA suffers from movable part identification and axis prediction as the number of parts increases, it leads to significant deviations in the motion axes. This in turn results in greater errors for part motion estimation and movable part reconstruction. The other two methods, ArtGS and ours, both achieve accurate reconstruction and motion estimation in many cases. However, for complex objects with adjacent movable parts, as reported in Table~\ref{tab:ours_cmp}, ArtGS leads to incorrect motion estimations and thus larger reconstruction errors, while our method constantly achieves robust estimation for all the movable parts.

Figure~\ref{fig:multi_obj} shows the reconstructed objects at different states and their motion structures. 
Comparing these multi-part articulated objects, ArtGS often fails and falls trapped in a local optimum early during optimization, since the clustering-based algorithm cannot provide a stable initialization.
By contrast, our over-segmentation-based initialization can segment potential part proposals from the object, and effectively merge them into complete parts during the adaptive integration optimization. Thus, our method produces robust reconstructions for all the multi-part objects.



In Figure~\ref{fig:merge_vis}, we show the intermediate results of movable parts being adaptively integrated under over-segmentation initialization.
For complex inputs, our method can merge similar parts based on motion parameters optimized after some iteration steps, resulting in complete movable parts with only a few cycles.






\begin{figure*}
  \centering
  \includegraphics[width=\linewidth]{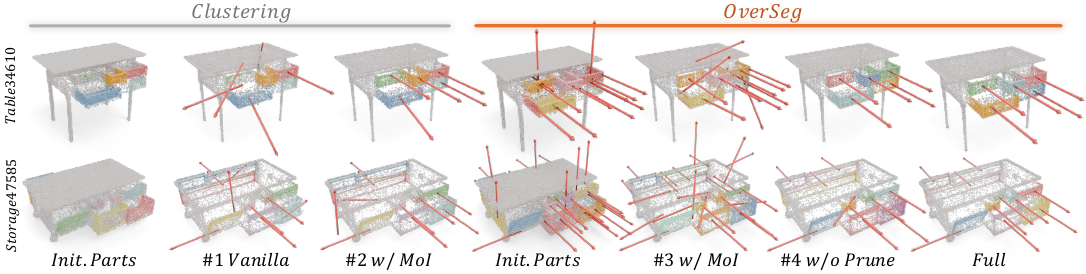}
  \caption{Ablation study results. To compare with our $OverSeg$ initialization, we use DBSCAN~\cite{ester1996density} as our clustering implementation, denoted $Vanilla$. We further add motion initialization to $Vanilla$ and $OverSeg$ baseline, denoted $w/\ MoI$. Finally, we removed the pruning strategy from $Full$ ($w/o\ Prune$) to verify its significant.}
  \label{fig:abstudy}
\end{figure*}
 
\begin{figure}[!ht]
  \centering
  \includegraphics[width=\linewidth]{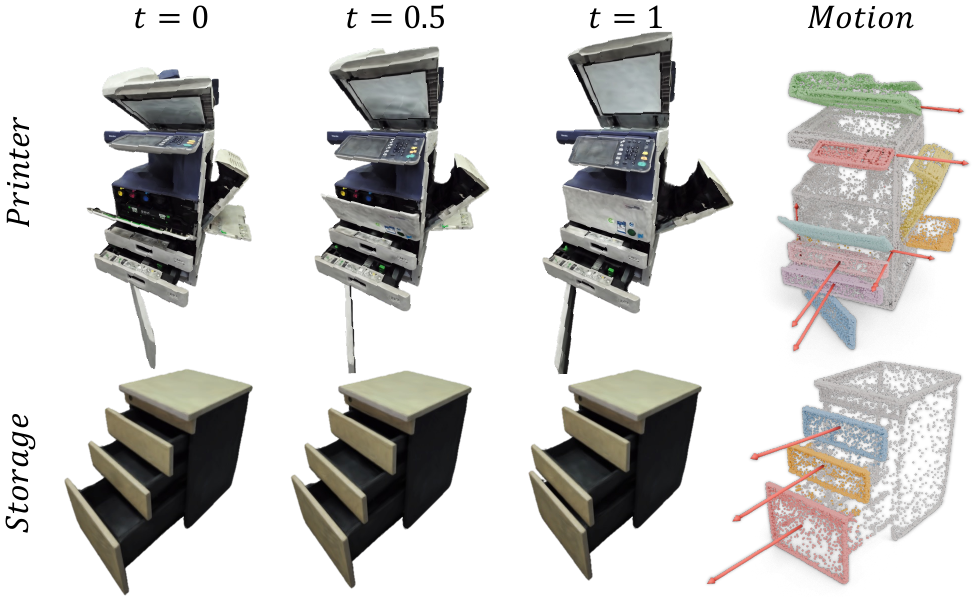}
  \caption{Reconstructed results of real-world articulated objects.}
  \label{fig:real_obj}
  \vspace{-3mm}
\end{figure}

\subsubsection{Results on Real-World Objects}

We additionally evaluate our approach on real-world articulated objects. Specifically, we collected 200 RGB images, 100 for each state, around the target object and utilized the pre-trained depth-anything-V2 model~\cite{yang2024depth} to estimate the calibrated depth. Then, we use SAM2~\cite{ravi2024sam2} to obtain the object masks. Finally, we take the processed multi-view RGBD images of two states as the input of our approach.

Figure~\ref{fig:real_obj} presents the reconstruction results of the real-world multi-part objects. Our method achieves robust and accurate results for them. More importantly, the superior mobility reconstruction ensures the correct articulated object with high-quality parts and motion parameters.



\subsection{Ablation Studies}

As listed in Table~\ref{tab:ablation}, we conduct the following ablation experiments to validate the effectiveness of our key designs:

\begin{itemize}
    \item The baseline (\#1) applies DBSCAN clustering~\cite{ester1996density} (with the ground-truth part number $M$) to separate the movable parts and assign identity transformation as initialization, then optimize the Gaussians and motion parameters with the same setting of our approach, but without the adaptive proposal integration and motion pruning strategy. As shown in Figure~\ref{fig:abstudy}, once the movable parts are incorrectly segmented or located close to each other, it sometimes causes missing parts or converges to local minima, leading to failed motion estimation and inaccurate geometry. 
    \item We then add our proposal initialization algorithm ($MoI$) into the baseline setting. Baseline (\#2) refers to initializing our proposals with the given part number $M$, while baseline (\#3) leverages our over-segmentation initialization without knowing the ground-truth part number. 
    In Figure~\ref{fig:abstudy}, the third col validates that our method leads to more reasonable mobility initialization and thus more accurate final reconstructions, while the fourth col exposes the remaining problem with unstable part initialization.
    \item We further insert our adaptive proposal integration strategy ($Merge$) in baseline (\#4). 
    It achieves robustness to the proposals of various complex articulated objects. However, some of the objects still exhibit relatively larger reconstruction errors, due to unstable motion initialization and the lack of collision-aware constraints to calibrate the parameters of adjacent parts, 
    such as the middle three drawers of Storage ($w/o\ Prune$) in Figure~\ref{fig:abstudy}.
    \item 
    By contrast, our full approach effectively prevents incorrect motion estimation caused by collisions between adjacent parts. By maximally leveraging motion priors of man-made objects, our method achieves robust reconstruction performance for all the multi-part objects.
\end{itemize}



\begin{table}
  \centering

  \caption{Quantitative evaluation of the ablation study. Lower($\downarrow$) is better on all metrics. To evaluate the results with arbitrary numbers of movable parts, we compute CD between the merged movable results and GT movable parts, denoted as over \overtext{CD}-m.}
  \resizebox{1.0\linewidth}{!}{
  \begin{tabular}{c| cccc| ccc}
    \toprule
    Case    & OverSeg & MoI & Merge & Prune & CD-s & \overtext{CD}-m & CD-w \\
    \midrule
    \#1     & &  & &
        	&	1.46	&	9.27	&	0.93 \\
    \#2     & & \checkmark & &
        	&	1.04	&	6.97	&	0.67 \\
    \#3     &   \checkmark  & \checkmark &  &
        	&	0.87	&	5.01	&	0.71 \\
    \#4     & \checkmark & \checkmark & \checkmark &
        	&	0.85	&	4.57	&	0.67 \\
    Full    & \checkmark & \checkmark & \checkmark & \checkmark
        	&	\textbf{0.84}	&	\textbf{3.63}	&	\textbf{0.65} \\
    \bottomrule
  \end{tabular}
  }
\vspace{-3mm}
  \label{tab:ablation}
\end{table}



    

\section{Conclusion}
We introduced ArtPro, a self-supervised framework based on 3D Gaussian Splatting for robustly reconstructing high-fidelity digital twins of articulated objects. Our method begins with an over-segmented set of part proposals and adaptively merges them during optimization through motion-consistency analysis and collision-aware pruning. This strategy significantly improves motion estimation and reconstruction quality, especially for complex objects with multiple parts and diverse articulations, as validated through extensive experiments on synthetic and real-world datasets.

However, our approach currently relies primarily on motion cues between two object states, which may be insufficient for disambiguating parts with perfectly symmetric or very subtle motions. Additionally, the reconstruction quality, particularly at part boundaries, remains sensitive to the accuracy of input sensor data. Future work will explore the integration of stronger semantic priors, geometry constraints, and multi-state tracking to address these challenges and further expand the applicability of our method.


\section{Acknowledgement}
This work is supported by the National Natural Science Foundation of China (62302269), the Excellent Young Scientists Fund Program (Overseas) of Shandong Province (No.2023HWYQ-034), the Joint Funds of the National Natural Science Foundation of China (U23A20312), a grant from the Chongqing Natural Science Foundation (CSTB2024NSCQ-MSX1026), the Natural Science Foundation of Shandong Province (No.ZR2023QF077), Shandong Province Key Research and Development Program (2024TSGC0051). We also acknowledge the support from the grant by Lejoin Intelligence (Shenzhen) Co.,Ltd.
{
    \small
    \bibliographystyle{ieeenat_fullname}
    \bibliography{main}
}

\clearpage
\setcounter{page}{1}
\maketitlesupplementary
\appendix

In this supplementary material, we first present the loss term formulations of Gaussian primitive optimization in Section~\ref{appendix:gs_opt}.
Then, we present more two-part comparisons in Section~\ref{appendix:more_paris}. In Section~\ref{appendix:more_multi} and Section~\ref{appendix:init_motion}, we provide more multi-part visualization and full initialization of our method on multi-part dataset.
In Section~\ref{appendix:add_ab} and Section~\ref{appendix:computation}, we report additional ablation studies and computational costs.
Finally, in Section~\ref{appendix:fail_case}, we report the failure cases of our method.

\section{Loss Terms of Gaussian Optimization}
\label{appendix:gs_opt}

Below we present the detailed formulation of the loss terms used in the Gaussian primitive optimization, which are mentioned in Section 3.3 and 3.4 in the main paper.

The RGBD loss $\mathcal{L}_{I}$ and the one-sided Chamfer Distance loss $\mathcal{L}_{cd}$ encourage the transformed Gaussians to align with the underlying surface of the end state. We have
\begin{equation}
\begin{aligned}
    \mathcal{L}_D=&\log(1+\parallel D_v(\mathcal{T}(\mathcal{G}))-\mathcal{D}_v\parallel_1)\\
    \mathcal{L}_{I}&=\mathcal{L}_{3dgs}(\mathcal{T}(\mathcal{G}))+\mathcal{L}_D(\mathcal{T}(\mathcal{G})),\\
    \mathcal{L}&_{cd}={\rm CD}(\cup\{x_m\}\rightarrow \mathcal P^1)
\end{aligned}
\end{equation}
where $D_v(\cdot)$ denotes the rendered depth map of view $v$, $\mathcal{L}_{3dgs}$ is the RGB loss used in 3DGS~\cite{Kerbl20233DGS}, $\mathcal{G}$ is the Gaussian sets, $\{x_m\}$ is the Gaussian centers of all movable $\{\mathcal{G}_m\}$, and $\mathcal{P}^1$ is the point cloud of end state. We compute the loss $\mathcal{L}_{I}$ between the rendered images of transformed Gaussian $\mathcal{T}(\mathcal{G})$ and GT RGBD of end state to optimize the appearance, and use $\mathcal{L}_{cd}$ to further improve the geometric quality.

The part contrastive loss $\mathcal{L}_{pc}$ ensures each Gaussian be dominated by one distinctive part
\begin{equation}
\mathcal{L}_{pc}=\mathbb{E}_{\{x_i\in\mathcal{G}_m\}^M}\Big[\frac{1}{M-1}\sum^M_{\substack{k=1 \\ k\ne m}}P_k(x_i)\Big],
\end{equation}
where, $\mathcal{G}_m$ denotes the $m$-th movable part of $\{\mathcal{G}_m\}^M_{m=1}$, $P_k(x_i)$ is the probability that the point $x_i$ belongs to the $k$-th movable part.

The local smoothness loss $\mathcal{L}_{ls}$ aims to maintain consistent static part probability for the spatial neighbors $x_i,x_j\in\mathcal{G}$. It is formulated as
\begin{equation}
\mathcal{L}_{ls}=\mathbb{E}_{x_i\in \mathcal{G},\ x_j\in{\rm kNN}(x_i,\mathcal{G})}| P_{s}(x_i)-P_{s}(x_j) |,
\end{equation}
where ${\rm kNN}$ denote k-nearest neighbors with $k=20$. To avoid fragmented parts, we smooth the static probability field on the kNN-based neighborhood graph to obtain a locally consistent segmentation field.

The regularization term aims to encourage the compactness of the parts. It uses the mean $\{\hat\mu_m\}$ and variance scaling $\{\hat s_m\}$ of initialization $\{\hat {\mathcal P}_m\}$ to maintain the existence of $p_m$
\begin{equation}
    \mathcal{L}_{reg}=\frac{1}{M}\sum_{m=1}^M\Big[\lambda_\mu | \mu_m-\hat\mu_m |+\lambda_s\parallel s_m-\hat s_m \parallel_1\Big]
\end{equation}
where, $s_m$ is the scaling of variance $\Sigma_m$, and $\lambda_\mu=0.5,\lambda_s=0.1$ is loss weight.

In summary, the objective function of Gaussian primitive optimization in the adaptive proposal integration stage is:
\begin{equation}
    \mathcal{L}=\mathcal{L}_{I}+\lambda_{cd}\mathcal{L}_{cd}+\lambda_{pc}\mathcal{L}_{pc}+\lambda_{ls}\mathcal{L}_{ls}+\lambda_{reg}\mathcal{L}_{reg}.
\end{equation}
We use $\lambda_{cd}=0.5,\lambda_{pc}=0.1,\lambda_{ls}=0.02$ and $\lambda_{reg}=1.0$ in all the experiments presented in this paper.



We further define a collision loss $\mathcal{L}_{col}$, which avoids the collisions between the movable and static parts of the transformed object. It is formulated as
\begin{equation}
\begin{aligned}
    &L_{col}(\mathcal{G}_i,\mathcal{G}_j)=\mathbb{E}_{x\in \mathcal{G}_i,y\in {\rm kNN}(x,\mathcal{G}_j)}|\hat\alpha_y|,\\
    \mathcal{L}_{col}&=L_{col}(\mathcal{T}(\mathcal{G}_m),\mathcal{G}_s)+L_{col}(\mathcal{G}_s,\mathcal{T}(\mathcal{G}_m)),
\end{aligned}
\end{equation}
where $\mathcal{T}(\mathcal{G}_m)$ is the union of all movable parts of $\mathcal{T}(\mathcal{G})$, $\mathcal{G}_s$ is the static part of $\mathcal{T}(\mathcal{G})$, $\hat\alpha_y$ is the Gaussian opacity corresponding to center position $y$, and ${\rm kNN}$ is the set of $k$ nearest neighbors with $k=32$.

Therefore, the objective function of the post-processing refinement optimization is 
\begin{equation}
\mathcal{L}=\mathcal{\mathcal{L}}_{I}(\mathcal{G})+\mathcal{\mathcal{L}}_{I}(\mathcal{T}(\mathcal{G}))+\lambda_c\mathcal{\mathcal{L}}_{col}
\end{equation}
We use $\lambda_c=0.02$ in all the experiments presented in this paper.


\begin{table*}[t] 
  \centering
  \footnotesize 
  \caption{Results on the PARIS~\cite{liu2023paris} dataset, including both synthetic and real data. Methods marked with an asterisk (*) denote versions with added depth supervision for fair comparison. Specifically, ArticulatedGS* and PARIS* are trained with an additional depth loss.}
   \label{tab:app_paris_cmp}
  \setlength{\tabcolsep}{3pt} 
  \resizebox{\linewidth}{!}{
  \begin{tabular}{ll ccccccccccc ccc}
    \toprule
    & & \multicolumn{11}{c}{Synthetic Objects} & \multicolumn{3}{c}{Real Objects} \\
    \cmidrule(lr){3-13} \cmidrule(lr){14-16}
    & & FoldChair & Fridge & Laptop & Oven & Scissor & Stapler & USB & Washer & Blade & Storage & All & Fridge & Storage & All \\
    \midrule
    
    \multirow{6}{*}{\makecell{Axis \\ Ang}} 
    & PARIS*~\cite{liu2023paris}
        &	15.79	&	2.93	&	0.03	&	7.43	&	16.62	&	8.17	&	0.71	&	0.71	&	41.28	&	0.03	&	9.37	&\Bs{1.90}	&	30.10	&	16.00	\\
    & ArticulatedGS~\cite{guo2025articulatedgs}
        &	6.20	&	0.14	&	15.55	&\Bs{0.00}	&\Ss{0.07}	&	0.08	&	0.16	&	0.03	&	0.23	&	0.04	&	2.25	&	4.06	&	48.56	&	26.31	\\
    & ArticulatedGS*
        &\Ss{0.02}	&	0.12	&	0.03	&	0.13	&	0.18	&	0.13	&	0.12	&	0.17	&\Ss{0.04}	&	0.04	&	0.10	&	64.81	&	29.44	&	47.12	\\
    & DTA~\cite{weng2024dta}
        &	0.03	&\Ss{0.09}	&	0.07	&	0.22	&	0.10	&	0.07	&\Ss{0.11}	&	0.36	&	0.20	&	0.09	&	0.13	&	2.08	&	13.64	&	7.86	\\
    & ArtGS~\cite{liu2025artgs}
        &\Bs{0.01}	&\Bs{0.03}	&\Ss{0.01}	&\Ss{0.01}	&\Bs{0.05}	&\Bs{0.01}	&\Bs{0.04}	&\Ss{0.02}	&\Bs{0.03}	&\Bs{0.01}	&\Bs{0.02}	&	2.09	&\Ss{3.47}	&\Ss{2.78}	\\
    & Ours
        &	0.04	&\Bs{0.03}	&\Bs{0.00}  &\Ss{0.01}	&\Bs{0.05}	&\Ss{0.04}	&\Bs{0.04}	&\Bs{0.01}	&	0.07	&\Ss{0.02}	&\Ss{0.03}	&\Ss{1.92}	&\Bs{1.02}	&\Bs{1.47}	\\
    \midrule

    \multirow{6}{*}{\makecell{Axis \\ Pos}} 
    & PARIS*~\cite{liu2023paris}
        &	0.25	&	1.13	&\Bs{0.00}	&	0.05	&	1.59	&	4.67	&	3.35	&	3.28	&	-	&	-	&	1.79	&	0.50	&	-	&	0.50	\\
    & ArticulatedGS~\cite{guo2025articulatedgs}
        &	4.93	&\Bs{0.00}	&	0.16	&	0.03	&\Bs{0.00}	&	0.02	&	0.63	&\Bs{0.00}	&	-	&	-	&	0.72	&	1.71	&	-	&	1.71	\\
    & ArticulatedGS*
        &\Bs{0.00}	&	0.02	&\Bs{0.00}	&	0.02	&\Bs{0.00}	&	0.02	&\Ss{0.64}	&\Ss{0.02}	&	-	&	-	&	0.09	&	2.71	&	-	&	2.71	\\
    & DTA~\cite{weng2024dta}
        &\Ss{0.01}	&\Ss{0.01}	&\Ss{0.01}	&\Ss{0.01}	&\Ss{0.02}	&	0.02	&\Bs{0.00}	&	0.05	&	-	&	-	&\Ss{0.02}	&	0.59	&	-	&	0.59	\\
    & ArtGS~\cite{liu2025artgs}
        &\Bs{0.00}	&\Bs{0.00}	&\Ss{0.01}	&\Bs{0.00}	&\Bs{0.00}	&\Ss{0.01}	&\Bs{0.00}	&\Bs{0.00}	&	-	&	-	&\Bs{0.00}	&\Ss{0.47}	&	-	&\Ss{0.47}	\\
    & Ours
        &\Bs{0.00}	&\Bs{0.00}	&\Bs{0.00}	&\Bs{0.00}	&\Bs{0.00}	&\Bs{0.00}	&\Bs{0.00}	&\Bs{0.00}	&	-	&	-	&\Bs{0.00}	&\Bs{0.34}	&	-	&\Bs{0.34}	\\
    \midrule

    \multirow{6}{*}{\makecell{Part \\ Motion}} 
    & PARIS*~\cite{liu2023paris}
        &	127.34	&	45.26	&	0.03	&	9.13	&	68.36	&	107.76	&	96.93	&	49.77	&	0.36	&\PS 0.30	&	50.52	&\PB 1.58	&	0.57	&	1.08	\\
    & ArticulatedGS~\cite{guo2025articulatedgs}
        &	59.08	&	0.55	&	28.16	&	0.13	&\PS 0.05	&	0.07	&	0.27	&\PB 0.00	&\PS 0.04	&\PB 0.00	&	8.84	&	15.11	&	0.53	&	7.82	\\
    & ArticulatedGS*
        &	0.26	&	0.12	&	0.07	&	0.31	&	0.13	&	0.09	&	0.10	&	0.09	&\PB 0.00	&\PB 0.00	&	0.12	&	39.58	&	0.32	&	19.95	\\
    & DTA~\cite{weng2024dta}
        &	0.10	&	0.12	&	0.11	&	0.12	&	0.37	&	0.08	&	0.15	&	0.28	&\PB 0.00	&\PB 0.00	&	0.13	&\PS 1.85	&\PS 0.14	&\PS 1.00	\\
    & ArtGS~\cite{liu2025artgs}
        &\PB 0.03	&\PS 0.04	&\PS 0.02	&\PB 0.02	&\PB 0.04	&\PB 0.01	&\PB 0.03	&\PS 0.03	&\PB 0.00	&\PB 0.00	&\PB 0.02	&	1.94	&\PB 0.04	&\PB 0.99	\\
    & Ours
        &\PS 0.06	&\PB 0.03	&\PB 0.00    &\PS 0.03	&\PB 0.04	&\PS 0.05	&\PS 0.04	&	0.04	&\PB 0.00	&\PB 0.00	&\PS 0.03	&	2.76	&\PB 0.04	&	1.40	\\
    \midrule

    \multirow{6}{*}{CD-s}
    & PARIS*~\cite{liu2023paris}
        &	10.20	&	8.82	&\PS 0.16	&	3.18	&	15.58	&	2.48	&	1.95	&	12.19	&	1.40	&	8.67	&	6.46	&	11.64	&	20.25	&	15.95	\\
    & ArticulatedGS~\cite{guo2025articulatedgs}
        &	3.17	&	2.02	&	4.27	&	2.31	&\PS 0.37	&	1.84	&	1.88	&	5.17	&	0.44	&	2.74	&	2.42	&	36.05	&	75.82	&	55.94	\\
    & ArticulatedGS*
        &	0.58	&	1.09	&	1.72	&\PS 1.88	&	0.64	&\PS 1.45	&\PS 1.24	&\PS 3.58	&\PS 0.24	&\PS 1.93	&\PS 1.44	&	230.04	&	46.78	&	138.41	\\
    & DTA~\cite{weng2024dta}
        &\PB 0.18	&	0.62	&	0.30	&	4.60	&	3.55	&	2.91	&	2.32	&	4.56	&	0.55	&	4.90	&	2.45	&	2.36	&	10.98	&	6.67	\\
    & ArtGS~\cite{liu2025artgs}
        &\PS 0.26	&\PS 0.52	&	0.63	&	3.88	&	0.61	&	3.83	&	2.25	&	6.43	&	0.54	&	7.31	&	2.63	&\PB 1.64	&\PS 2.93	&\PS 2.29	\\
    & Ours
        &	0.48	&\PB 0.31	&\PB 0.12	&\PB 1.41   &\PB 0.19	&\PB 0.67	&\PB 0.96	&\PB 2.69	&\PB 0.21	&\PB 1.81	&\PB 0.89	&\PS 1.96	&\PB 2.54	&\PB 2.25	\\
    \midrule

    \multirow{6}{*}{CD-m}
    & PARIS*~\cite{liu2023paris}
        &	17.97	&	7.23	&	0.15	&	6.54	&	16.65	&	30.46	&	10.17	&	265.27	&	117.99	&	52.34	&	52.48	&	77.85	&	474.57	&	276.21	\\
    & ArticulatedGS~\cite{guo2025articulatedgs}
        &	36.61	&	2.29	&	24.90	&	0.96	&\PS 0.35	&	1.64	&\PS 1.02	&	3.88	&	1.86	&	5.49	&	7.90	&	107.96	&	2,459.45	&	1,283.71	\\
    & ArticulatedGS*
        &	0.33	&	0.63	&	2.07	&	1.08	&	0.57	&	1.97	&	1.17	&\PS 0.41	&\PB 0.78	&	0.83	&\PS 0.98	&	69.23	&	1,578.02	&	823.63	\\
    & DTA~\cite{weng2024dta}
        &\PS 0.15	&\PS 0.27	&\PS 0.13	&\PS 0.44	&	10.11	&	1.13	&	1.47	&	0.45	&	2.05	&\PS 0.36	&	1.66	&\PS 1.12	&\PS 30.78	&\PS 15.95	\\
    & ArtGS~\cite{liu2025artgs}
        &	0.54	&\PB 0.21	&\PS 0.13	&	0.89	&	0.64	&\PB 0.52	&	1.22	&	0.45	&\PS 1.12	&	1.02	&\PB 0.67	&\PB 0.66	&\PB 6.28	&\PB 3.47	\\
    & Ours
        &\PB 0.11	&	0.28	&\PB 0.07	&\PB 0.33	&\PB 0.18	&\PS 0.83	&\PB 0.37	&\PB 0.08	&	F  	&\PB 0.35	&	0.28$\ ^\text{F}$&	28.31	&	35.95	&	32.13	\\
    \midrule

    \multirow{6}{*}{CD-w}
    & PARIS*~\cite{liu2023paris}
        &	4.37	&	5.53	&\PS 0.26	&	3.18	&	3.90	&	5.27	&	1.78	&	10.11	&	0.58	&	7.80	&	4.28	&	8.99	&	32.10	&	20.55	\\
    & ArticulatedGS~\cite{guo2025articulatedgs}
        &	1.06	&	2.12	&	8.52	&	2.13	&\PS 0.35	&	1.61	&	1.88	&	4.79	&	0.23	&	2.62	&	2.53	&	77.53	&	995.99	&	536.76	\\
    & ArticulatedGS*
        &	0.43	&	0.96	&	0.80	&\PS 1.60	&	0.59	&\PS 1.48	&\PS 1.00	&\PS 3.08	&\PS 0.20	&\PS 1.75	&\PS 1.19	&	51.22	&	965.80	&	508.51	\\
    & DTA~\cite{weng2024dta}
        &\PS 0.27	&	0.70	&	0.32	&	4.24	&	0.41	&	1.92	&	1.17	&	4.48	&	0.36	&	3.99	&	1.79	&	2.08	&	8.98	&	5.53	\\
    & ArtGS~\cite{liu2025artgs}
        &	0.43	&\PS 0.58	&	0.50	&	3.58	&	0.67	&	2.63	&	1.28	&	5.99	&	0.61	&	5.21	&	2.15	&\PS 1.29	&\PS 3.23	&\PS 2.26	\\
    & Ours
        &\PB 0.10	&\PB 0.31	&\PB 0.09	&\PB 1.32	&\PB 0.18	&\PB 0.62	&\PB 0.50	&\PB 2.40	&\PB 0.18	&\PB 1.60	&\PB 0.73	&\PB 1.03	&\PB 2.03	&\PB 1.53	\\

    \bottomrule
  \end{tabular}
  }
  
\end{table*}


\begin{figure*}[!ht]
  \centering
  \includegraphics[width=\linewidth]{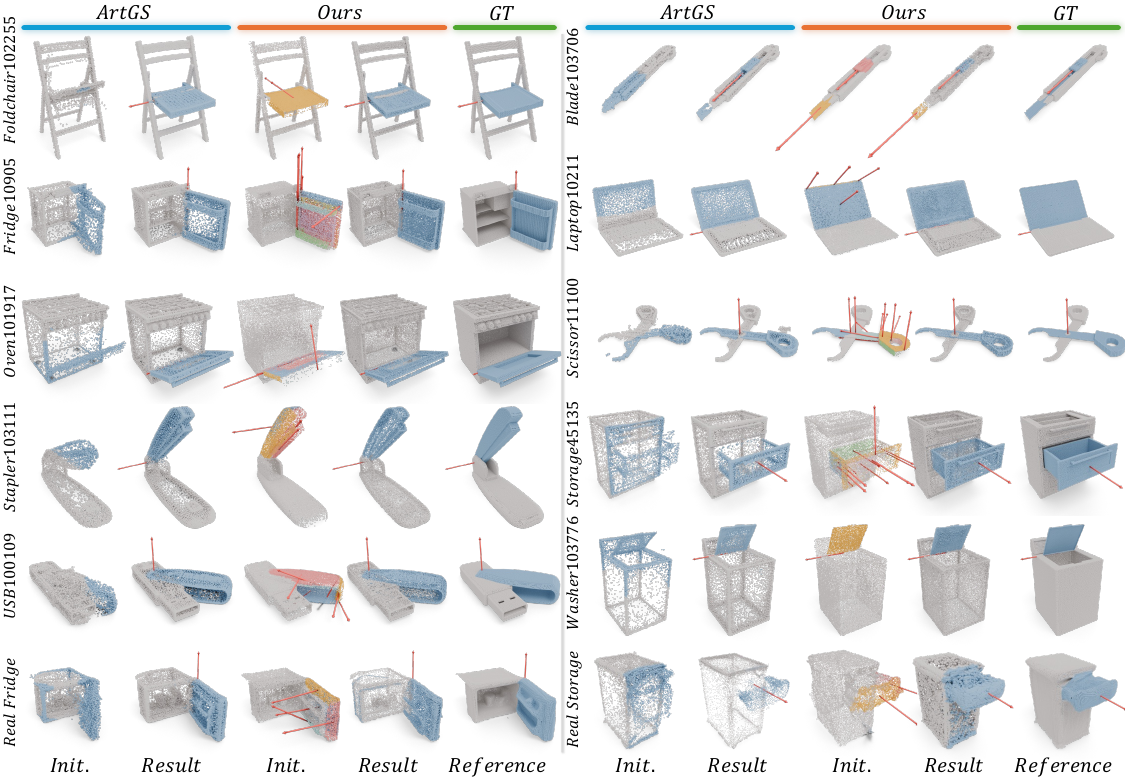}
  \caption{Additional qualitative results on PARIS dataset, including the initializations and the final reconstructions. We show the Gaussians with their center points for a better visualization of their segmentation and motion parameters.}
  \label{fig:app_more_paris}
\end{figure*}
 
\begin{figure*}[!ht]
  \centering
  \includegraphics[width=\linewidth]{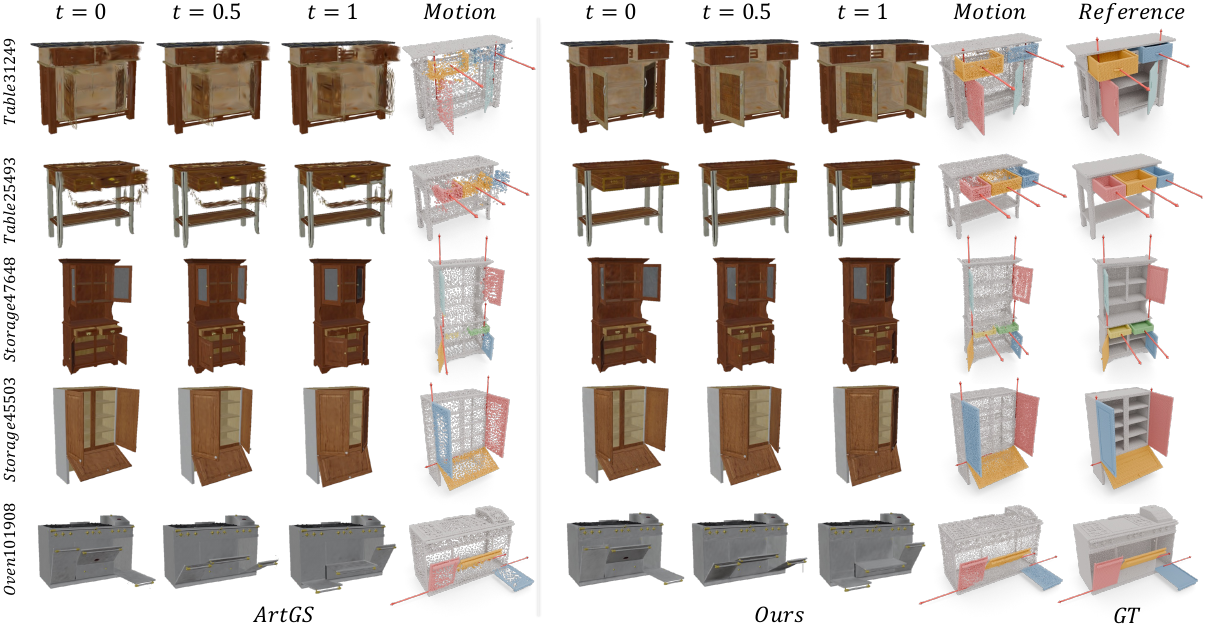}
  \caption{Additional reconstruction results on \textbf{ArtGS-Multi dataset}. We render the reconstructed articulated objects in different motion states ($t \in \{0,0.5,1\}$ and their motion structures for each result.}
  \label{fig:app_more_artgs}
\end{figure*}
 
\begin{figure*}[!ht]
  \centering
  \includegraphics[width=\linewidth]{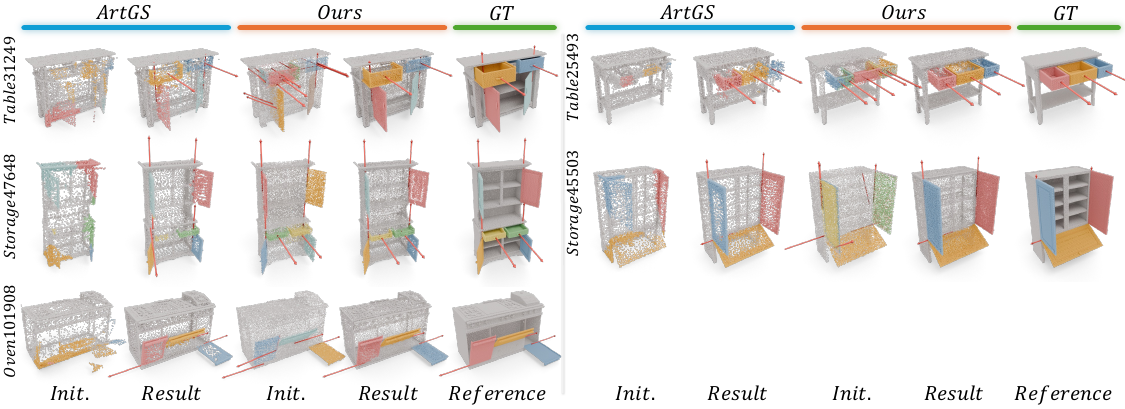}
  \caption{Additional qualitative results on \textbf{ArtGS-Multi dataset}, including the initializations and the final reconstructions. We show the Gaussians with their center points for a better visualization of their segmentation and motion parameters.}
  \label{fig:app_more_artgs_i}
\end{figure*}
 
\begin{figure*}[!ht]
  \centering
  \includegraphics[width=\linewidth]{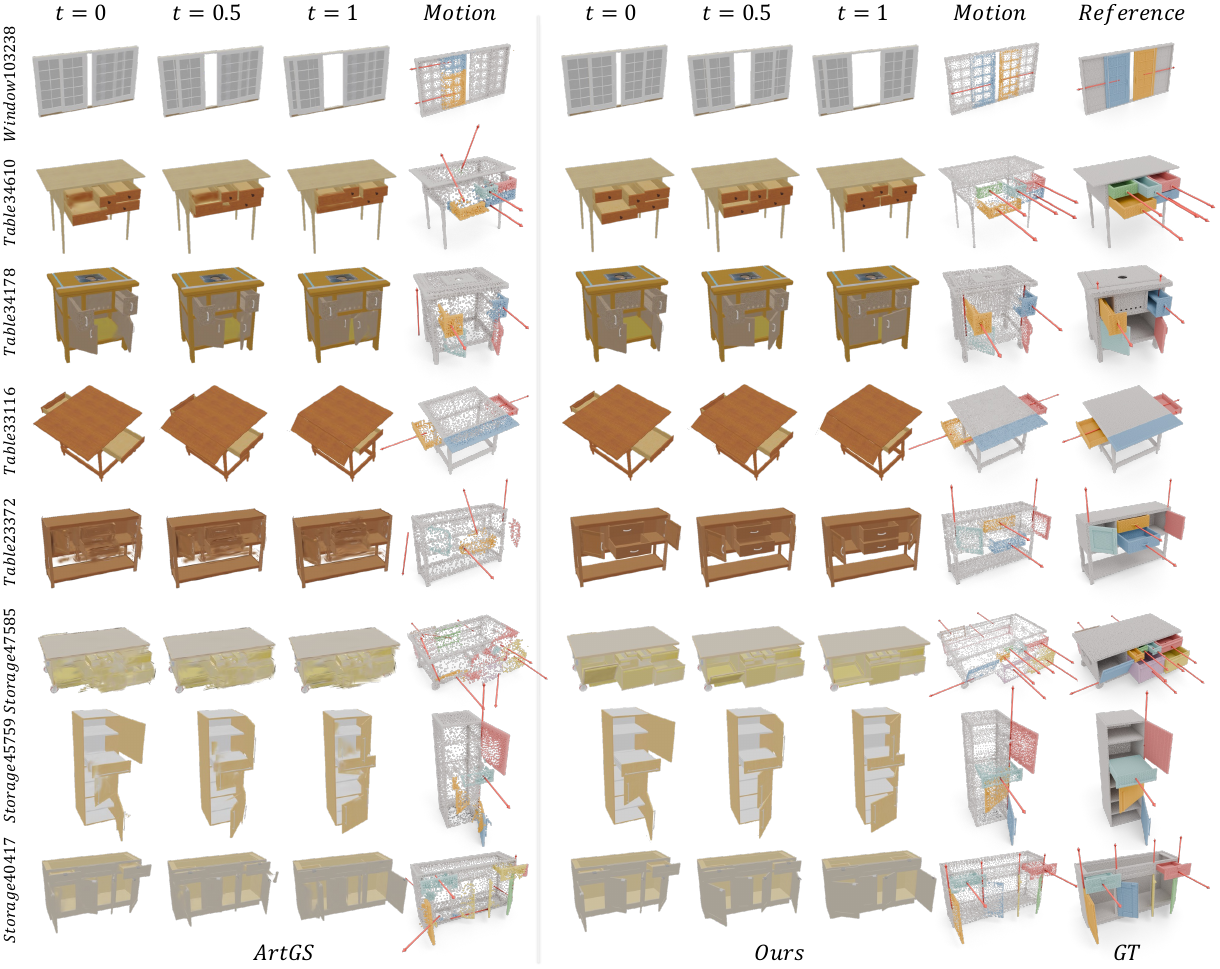}
  \caption{Additional reconstruction results on \textbf{our dataset}. We render the reconstructed articulated objects in different motion states ($t \in \{0,0.5,1\}$ and their motion structures for each result.}
  \label{fig:app_more_ours}
\end{figure*}
 
\begin{figure*}[!ht]
  \centering
  \includegraphics[width=\linewidth]{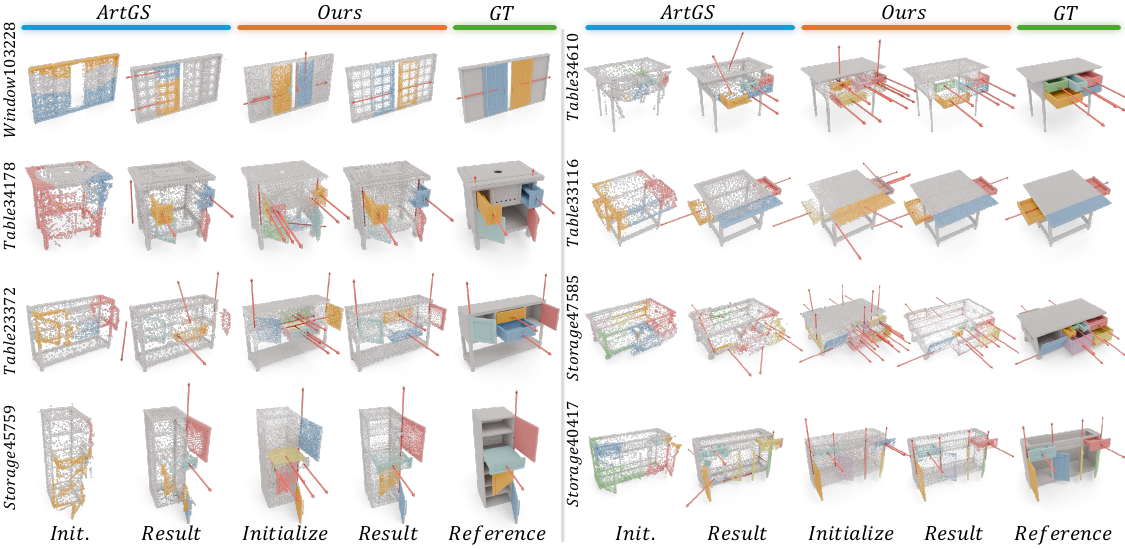}
  \caption{Additional qualitative results on \textbf{our dataset}, including the initializations and the final reconstructions. We show the Gaussians with their center points for a better visualization of their segmentation and motion parameters.}
  \label{fig:app_more_ours_i}
\end{figure*}

\section{Additional Results on Two-Part Dataset}
\label{appendix:more_paris}

We compare with related methods including PARIS~\cite{liu2023paris}, ArticulatedGS~\cite{guo2025articulatedgs}, DTA~\cite{weng2024dta}, ArtGS~\cite{liu2025artgs} on the two-part object dataset. Since we take RGBD images as input, we apply the depth loss to PARIS~\cite{liu2023paris} and ArticulatedGS~\cite{guo2025articulatedgs} methods for a fair comparison.
We report the quantitative evaluation results in Table~\ref{tab:app_paris_cmp}, showing that the 3DGS-based methods, i.e. ArticulatedGS~\cite{guo2025articulatedgs}, ArtGS~\cite{liu2025artgs}, and ours,  achieve competitive performance. This is because static and movable parts can be well initialized with spatial clustering for two-part objects, which are further refined with the following deformable 3DGS optimization. This experiment also acts as a sanity check for our method, validating that the over-segmentation proposals from the same movable part can be effectively merged based on their motion similarities.

Figure~\ref{fig:app_more_paris} provides the full visualization results of ArtGS and Ours on this dataset. The results demonstrate that our method can robustly integrate the proposals and reconstruct both the part probability field and motion parameters. For incorrectly initialized mobility proposals, our optimization can also obtain correct motion estimation and the final reconstructions.
For the real-world objects in this dataset, our method's performance on the geometry reconstruction, especially in terms of CD metric, is limited by the quality of input depth maps. This is because our method tends to estimate the mobility parameters that align the transformed results closer to the depth map of end state ($t=1$). Large discrepancies between the input depth maps of states $t=0$ and $t=1$ lead to our results containing a small number of Gaussians originating from static part, such as the Real Storage in Figure~\ref{fig:app_more_paris} and Figure~\ref{fig:app_fail_case}.

\section{Additional Results on Multi-Part Dataset}
\label{appendix:more_multi}

We present a comprehensive visual analysis to compare ArtGS~\cite{liu2025artgs} and our method in Figure~\ref{fig:app_more_artgs} and Figure~\ref{fig:app_more_ours}, which show the reconstructed articulated objects at different motion states and their articulated joints of movable parts. Compared to ArtGS, our method can robustly estimate the movable parts and motion parameters, leading to consistent and clean rendered images of the reconstructed objects.

\section{More Visual Analysis on Multi-Part Dataset}
\label{appendix:init_motion}

We further show the visualizations of the initializations and reconstructed results of ArtGS and ours in Figure~\ref{fig:app_more_artgs_i} and Figure~\ref{fig:app_more_ours_i}.
The results show that our mobility initialization can effectively extract the potential movable parts and reasonable motion initialization, while ArtGS suffers from unstable initialization. For over-segmented proposals, our optimization can effectively estimate their motion parameters and integrate adjacent proposals into an individual part.

We further report the visual quality metrics on our dataset in Table~\ref{tab:visual_quality}.
Our method achieves better PSNR, SSIM, and LPIPS scores compared to ArtGS, while producing more accurate articulation.

\begin{table}[ht]
\vspace{-10pt}
\centering
    \caption{Visual quality on our dataset.}
    \vspace{-5pt}
    \begin{tabular}{l|ccc}
      \toprule
      Method & PSNR  & SSIM  & LPIPS \\
      \midrule
      ArtGS  & 36.10 & 0.980 & 0.037 \\
      Ours   & \textbf{48.02} & \textbf{0.998} & \textbf{0.003} \\
      \bottomrule
    \end{tabular}
\vspace{-10pt}   
\label{tab:visual_quality}
\end{table}

\section{More Ablation Study Results}
\label{appendix:add_ab}

To strengthen the validation of careful initialization, we report the results of ``w/o MoI" (no principal axes for initialization) and ``w/o MoI$\&$PI" (no principal axes for initialization and pruning) in Table~\ref{tab:more_ablation}. We also add the results of disabling the progressive merging after over-segmentation (denoted as w/o PM) in Table~\ref{tab:more_ablation}.

\begin{table}
\vspace{-10pt}
    \centering
    \caption{Additional ablation study on our dataset.}
    \vspace{-5pt}
    \begin{tabular}{c|ccc}
      \toprule
      Case     & CD-s & \overtext{CD}-m & CD-w \\
      \midrule
      w/o\ PM
            &	0.85	&	4.26	&	0.68 \\
      w/o\ MoI    
          	&	0.89	&	5.62	&	0.70 \\
      w/o\ MoI\&PI    
          	&	0.87	&	5.81	&	0.71 \\
      Full    
          	&	\textbf{0.84}	&	\textbf{3.63}	&	\textbf{0.65} \\
      \bottomrule
    \end{tabular}
    \vspace{-10pt}
  \label{tab:more_ablation}
\end{table}

\begin{table}[ht]
\centering
\caption{Computation overhead comparison on our multi-part dataset. Time and memory are reported as mean${\scriptstyle\pm}$std.}
\begin{tabular}{lcc}
\toprule
Method & Time (min) & GPU Memory (GB) \\
\midrule
DTA~\cite{weng2024dta}    & 53.28${\scriptstyle\pm23.78}$ & 10.58${\scriptstyle\pm2.45}$ \\
ArtGS~\cite{liu2025artgs} & 10.41${\scriptstyle\pm0.89}$  &  2.94${\scriptstyle\pm0.04}$ \\
Ours                       & 16.93${\scriptstyle\pm2.37}$  &  2.69${\scriptstyle\pm0.23}$\textsuperscript{*} \\
\bottomrule
\end{tabular}
\vspace{-10pt}
\label{tab:time_and_memory}
\end{table}

\section{Computation Overhead}
\label{appendix:computation}

We control the computational cost of ArtPro through two key strategies.
First, we use PartField~\cite{liu2025partfield} to produce well-structured over-segmentation proposals rather than fragmented patches, and tend to merge the most relevant proposals in the first optimization cycle, which limits the total number of proposals and cycles.
Second, the optimization of each cycle is monitored and stops early upon convergence.

Table~\ref{tab:time_and_memory} reports the average computation time and GPU memory usage across all objects in our dataset.
Our method achieves a practical balance between computational cost and reconstruction quality. Although ArtGS is faster, it cannot robustly reconstruct complex multi-part objects, as shown in Tables~1--2 of the mainpage. DTA, on the other hand, requires substantially more time and memory while also failing on many multi-part cases.

\begin{figure*}[!ht]
  \centering
  \includegraphics[width=\linewidth]{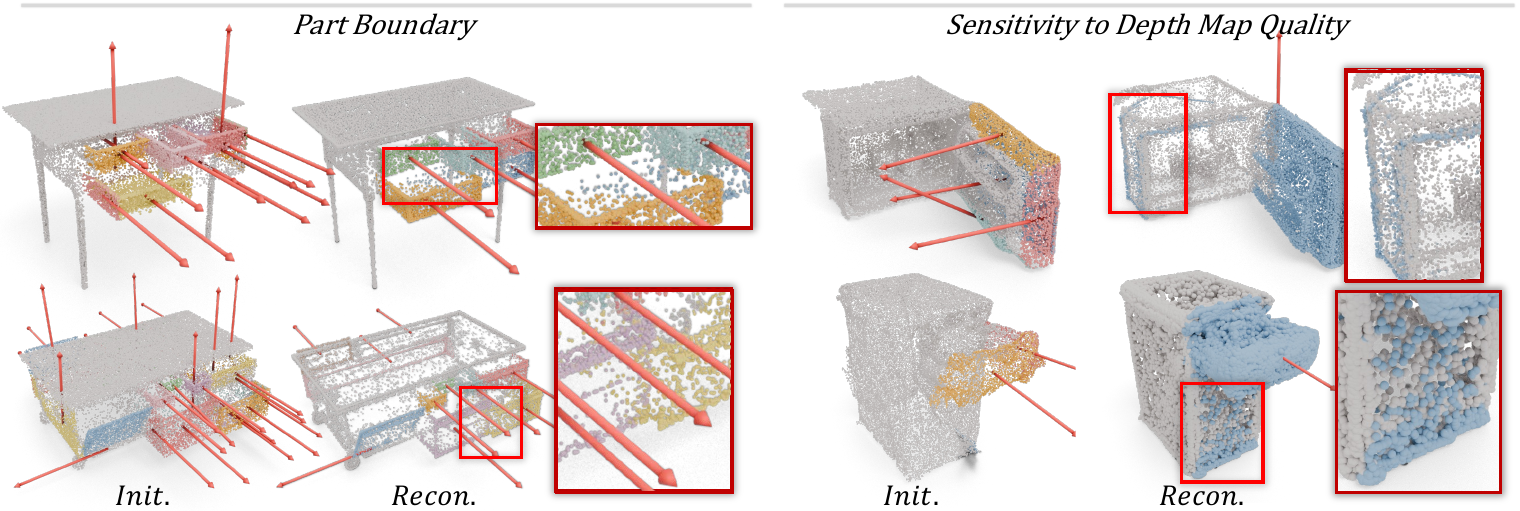}
  \caption{Failure cases. We illustrate failure cases of low-quality part boundaries. For low-quality RGBD images input, our method difficult to reconstruct the part boundaries.}
  \label{fig:app_fail_case}
\end{figure*}
 
\begin{figure}[!t]
  \centering
  \includegraphics[width=\linewidth]{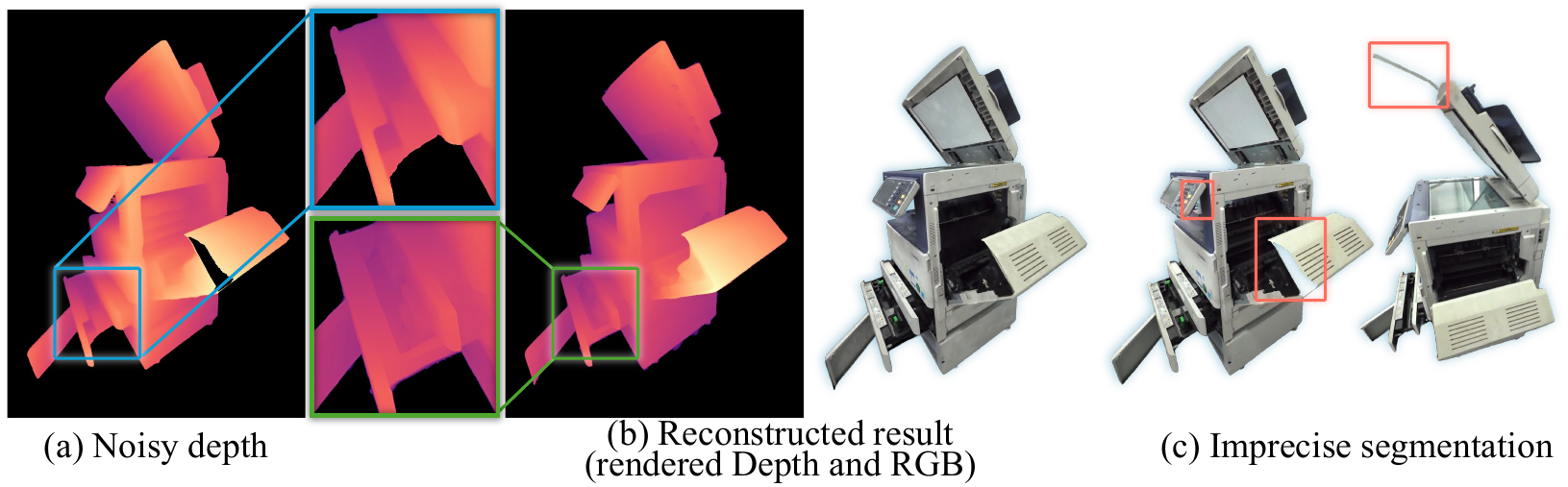}
  \vspace{-21pt}  
  \caption{Influence of noisy depth and mask.}
  \label{fig:noisy_depth_sam}
  \vspace{-10pt}
\end{figure}

\section{Failure Cases and Robustness Analysis}
\label{appendix:fail_case}

\noindent\textbf{Robustness to noisy depth and masks.}
Fig.~\ref{fig:noisy_depth_sam} shows the robustness of our method to noisy depth maps and imprecise segmentation masks.
For local geometry errors (Fig.~\ref{fig:noisy_depth_sam}(a)) and imprecise mask boundaries near drawer edges (Fig.~\ref{fig:noisy_depth_sam}(c)), our method mitigates the incorrect estimation from specific views by integrating multi-view information, achieving accurate reconstructions.
However, our method would fail when the captured depth maps and masks are highly fragmented and misaligned, as in the real-scanned data provided by PARIS~\cite{liu2023paris}.
This can be enhanced using prior-prompted depth predictor, e.g., PriorDA~\cite{wang2025depth}, as we did to obtain our real-scanned data (the printer and storage).

\noindent\textbf{Part boundary sensitivity.}
Although our approach achieves much more robust motion estimation and reconstruction of articulated objects than existing approaches, it still suffers from the common issues of the self-supervised 3DGS reconstruction framework. We show two representative failure cases in Figure~\ref{fig:app_fail_case}. First, since our method lacks semantic part segmentation supervision for separating the adjacent movable parts, it sometimes causes the part boundary to encroach on neighboring part regions, such as the interior and the edge of the drawers in Figure~\ref{fig:app_fail_case}. 

\noindent\textbf{Sensitivity to depth map quality.}
Second, the estimation of mobility parameters in our method primarily relies on the depth component of the RGBD loss $\mathcal{L}_{I}$ and the geometry-aware Chamfer distance loss $\mathcal{L}_{cd}$. Consequently, the accuracy of these estimations is highly dependent on the input depth map quality. In scenarios where the acquired depth maps contain significant inaccuracies, due to factors such as sensor noise, occlusions, or varying illumination conditions, the optimization process can be misled. As illustrated in Figure~\ref{fig:app_fail_case} (using real-world data from the PARIS dataset), this can cause our method to erroneously incorporate Gaussians from the static part into the estimated movable part. To achieve a lower loss value under these incorrect constraints, the optimization incorrectly transforms the movable part into the interior of the object. It is worth noting that in our real-world reconstruction experiments, the quality of the mobility results has been significantly improved by enhancing the input depth maps using the pre-trained Depth-Anything-V2 model~\cite{yang2024depth} which effectively mitigates these issues.

\noindent\textbf{Multi-DOF joints and non-rigid parts.}
Our current formulation assumes that each movable part undergoes a single rigid transformation, either revolute or prismatic.
This assumption does not hold for objects with multi-DOF joints such as ball joints or compound hinges, or parts exhibiting non-rigid deformations.
In such cases, our motion parameterization cannot capture the full range of part motion, leading to inaccurate articulation estimation.
Future work will explore potential solutions including extending the motion model to support composite transformations or integrating learned deformation fields for non-rigid components.

\end{document}